\pdfoutput=1

\documentclass[11pt,table]{article}

\usepackage[]{acl}

\usepackage[]{xcolor}
\usepackage[T1]{fontenc} 
\usepackage{fancyvrb}
\usepackage{tcolorbox}
\usepackage{makecell}
\usepackage{geometry}

\usepackage{times}
\usepackage{latexsym}
\usepackage{graphicx}
\usepackage{xspace}
\usepackage{booktabs}
\usepackage{amsmath,amssymb}
\usepackage{tcolorbox}
\usepackage{enumitem}

\tcbset{
  promptbox/.style  = {
    enhanced,
    sharp corners,
    colback=white,
    colframe=black,
    boxrule=0.5pt,
    fonttitle=\bfseries,
    colbacktitle=black,
    coltitle=white,
    boxed title style={sharp corners},
    attach title to upper,
    left=1mm, right=1mm, top=1mm, bottom=1mm,
    before skip=1em, after skip=1em,
    width=\columnwidth,
  }
}

\usepackage[T1]{fontenc}

\usepackage[utf8]{inputenc}

\usepackage{microtype}

\usepackage{inconsolata}

\usepackage{graphicx}

\newcommand{\DialogXpert}{\texttt{DialogXpert}\xspace}
\newcommand{\lt}{<}
\newcommand{\gt}{>}
\title{\DialogXpert: Driving Intelligent and Emotion-Aware Conversations through Online Value-Based Reinforcement Learning with LLM Priors}

\author{
    \textbf{Tazeek Bin Abdur Rakib\textsuperscript{1}},
    \textbf{Ambuj Mehrish\textsuperscript{2}}\\
    \textbf{Lay-Ki Soon\textsuperscript{1}},
    \textbf{Wern Han Lim\textsuperscript{1}},
    \textbf{Soujanya Poria\textsuperscript{2}}\\
    \textsuperscript{1}School of Information Technology, Monash University Malaysia\\
    \textsuperscript{2}Singapore University of Technology and Design\\
    \texttt{\{soon.layki, lim.wern.han, tazeek.binabdurrakib\}@monash.edu}\\
    \texttt{\{ambuj\_mehrish, sporia\}@sutd.edu.sg}
  }

\begin{document}
\maketitle
\begin{abstract}
Large-language-model (LLM) agents excel at reactive dialogue but struggle with proactive, goal-driven interactions due to myopic decoding and costly planning. We introduce \DialogXpert, which leverages a frozen LLM to propose a small, high-quality set of candidate actions per turn and employs a compact Q-network over fixed BERT embeddings trained via temporal-difference learning to select optimal moves within this reduced space. By tracking the user's emotions, \DialogXpert tailors each decision to advance the task while nurturing a genuine, empathetic connection. Across negotiation, emotional support, and tutoring benchmarks, \DialogXpert drives conversations to under $3$ turns with success rates exceeding 94\% and, with a larger LLM prior, pushes success above 97\% while markedly improving negotiation outcomes. This framework delivers real-time, strategic, and emotionally intelligent dialogue planning at scale\footnote{Code available at \url{https://github.com/declare-lab/dialogxpert/}
}.
\end{abstract}

\section{Introduction}

Recent advances in large language models (LLMs) such as ChatGPT \citep{openai2022chatgpt}, Vicuna \cite{zheng2023judging}, and LLaMA2-Chat \citep{Ouyang2022TrainingLM,Touvron2023LLaMAOA} have significantly enhanced open-domain dialogue systems, enabling fluent, context-aware, and intent-aligned responses \cite{hu2023enhancing}. However, these systems remain largely reactive, adept at replying to user input but limited in proactively steering conversations toward specific goals. Domains such as negotiation, emotional support, and tutoring require initiative and long-term planning \cite{deng2023prompting, kang2024can, song2024typing}, which current LLMs often lack \cite{deng2025proactive}.

This limitation stems from their turn-by-turn generation, typically guided by greedy decoding, that overlooks future dialogue objectives \cite{levin1997learning, emnlp22-esc}. Although techniques like Monte Carlo Tree Search (MCTS) \cite{44806, zhao2024probabilistic} and $A^*$ search \citep{hart1968formal} offer deeper look-ahead \cite{vath2023conversational}, they are computationally expensive and unsuitable for real-time use.

Prior to LLMs, dialogue planning relied on supervised learning over annotated corpora \citep{iclr20-noncollab,iclr21-negotiate-strategy,emnlp22-esc,eacl23-strategy-tutor,kemi,pacific}, focusing on dialogue act prediction. These approaches were static, domain-specific, and difficult to scale, often failing to adapt to evolving user behavior or optimize long-term outcomes. While LLMs introduced a new paradigm, efficient and goal-driven dialogue planning remains an open challenge.

To mitigate these challenges, recent frameworks such as Plug-and-Play Dialogue Policy Planning (PPDPP)~\citep{deng2024plug} have emerged. PPDPP fine-tunes a compact RoBERTa-based~\citep{roberta} policy language model using supervised learning and further optimizes it through self-play~\citep{Silver2017MasteringCA} with LLM-based user and reward simulators. This approach is computationally efficient requiring only a single forward pass per turn but remains inherently myopic. It selects actions greedily, lacks multi-turn foresight, and is constrained by the limited zero- or few-shot generalization capabilities of the frozen policy model. Consequently, the agent may choose locally optimal but globally suboptimal actions and struggle with out-of-distribution states.

Dual-Process Dialogue Planner (DPDP) \citep{he2024planning} improves over PPDPP with Kahneman’s dual-process theory \citep{kahneman2003mapsob}, pairing a fast RoBERTa policy (System 1) with an MCTS planner (System 2) triggered under uncertainty \citep{Anthony2017ThinkingFA}. While this boosts look-ahead reasoning, repeated rollouts and reward simulations incur high latency, and its heuristic gating can misjudge when deeper reasoning is needed. Moreover, both DPDP and PPDPP rely on compact, fine-tuned models that either plan too greedily or at excessive computational cost.

We propose the \textit{LLM-Prior Planning Paradigm}, which leverages frozen LLMs’ generalization without full-tree planning overhead. At each turn, a frozen LLM (e.g., Qwen-2.5 14B \citep{bai2023qwen}) produces a top-$k$ set of semantically coherent actions, forming a concise prior \citep{Bengio2017TheCP, korbak2022rl}. A lightweight \textit{Q-network, trained via Q-learning on fixed BERT embeddings of state–action pairs} \citep{devlin2018bert, mnih2013playing}, performs localized rollouts within this candidate set and updates value estimates through temporal-difference learning \citep{watkins1992q, tesauro1995temporal, yan2024efficient}. This reduces expensive LLM calls, avoids exhaustive tree expansion, and converges rapidly even in compact action spaces.

Importantly, dialogue effectiveness depends not only on task success but also on emotional resonance \citep{chen2023effective, asghar2020generating}. To this end, we introduce \DialogXpert, an LLM-Prior framework enhanced with a dedicated \textit{emotion-tracking component}. After each system turn, \textit{the Emotion Tracker} infers the user’s current feelings for example, distress or engagement from the chosen action and preceding context. These inferred emotions are folded into the planner’s state representation, allowing \DialogXpert to trade off goal progress against rapport building. As a result, the agent avoids abrupt or tone-deaf responses, producing conversations that feel both effective and genuinely empathetic \citep{Zhao2023IsCE}.

\begin{figure*}[t]
  \centering
  \includegraphics[width=\textwidth]{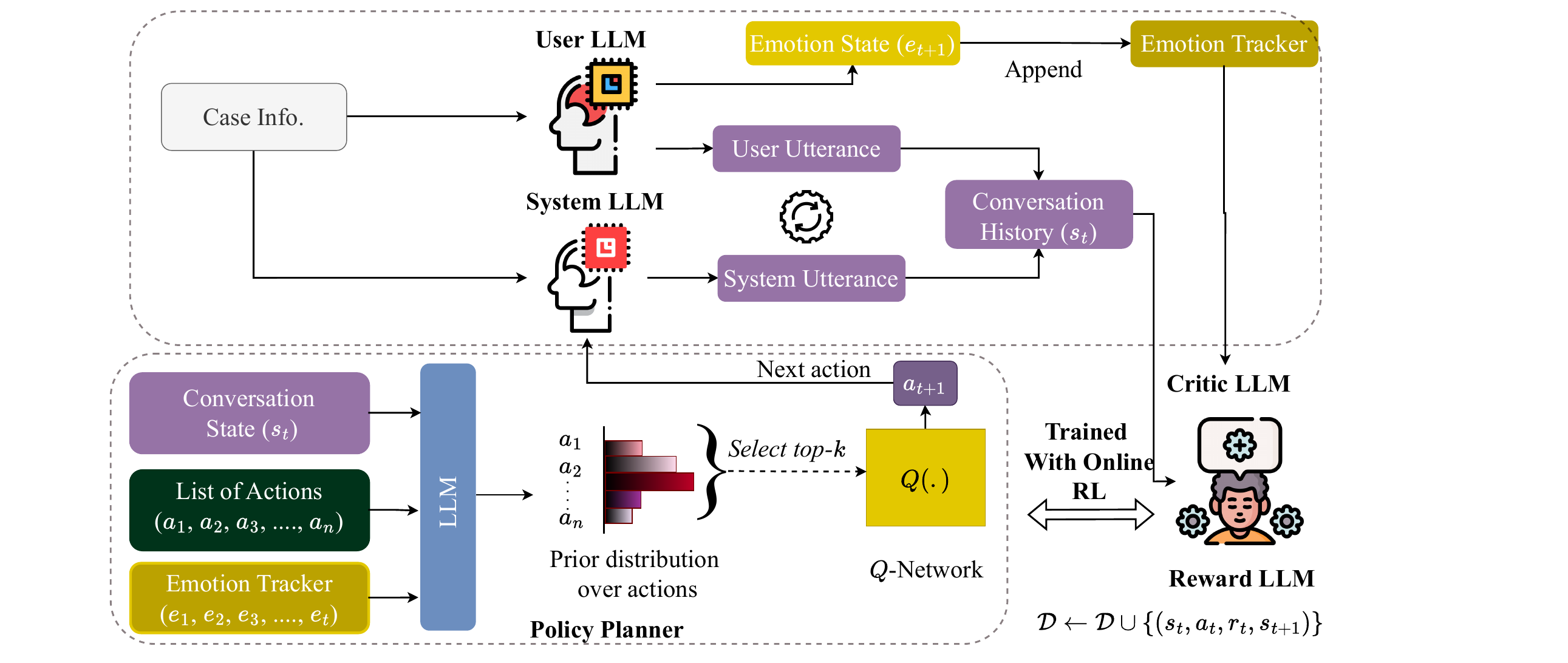}
  \caption{ \DialogXpert pipeline: case information and dialogue history drive user/system LLMs and an emotion tracker; a frozen LLM generates a prior over candidate actions, the top-k are evaluated by a Q-network and executed by the system LLM; a critic LLM provides reward signals to train the Q-network.}
  \label{Fig:archi}
\end{figure*}

Our contributions are: (1) \DialogXpert model that combines the strategic power of LLMs \cite{Xu2023SmallMA} with the efficiency of lightweight value learning and the sensitivity of emotion-aware planning. (2) It tackles major limitations seen in earlier approaches like short-sighted decisions, poor generalization, and heavy computational demands while still being suitable for real-time use. (3) Results across a range of tasks, including negotiation, tutoring, and emotional support, demonstrate its strong performance, setting a new standard for proactive and emotionally intelligent dialogue systems.
\vspace{-0.5em}
\section{Related Works}
LLM-driven decision-making has progressed from fine-tuned chatbots to sophisticated planners. Early systems like DialoGPT \citep{zhang2019dialogpt,Zhang2020LearningGD}, ProAgent \citep{zhang2023proagent}, and Voyager \citep{wang2023voyager} adapted pretrained transformers or retrieval-augmented controllers for multi-step tasks, while prompt-chaining (Proactive, ProCoT \citep{deng2023prompting}) and modular prompting (Ask-an-Expert \citep{zhang2023ask}, ICL-AIF \citep{fu2023improving}) enabled iterative reasoning and decomposed tasks. Planning-as-search methods such as Tree-of-Thoughts \citep{yao2023tree}, RAP with MCTS rollouts \citep{hao2023reasoning}, and reinforcement learning approaches like PPDPP \citep{deng2024plug} and DPDP \citep{he2024planning} improved exploration efficiency. Recent latent-policy techniques such as LDPP \citep{he2025simulation} and UDP \citep{he2025simulating} learn continuous action representations via VAE and diffusion-based user models. 
In contrast, \DialogXpert treats the LLM as a frozen action proposer: it selects top-$k$ samples from a large pretrained model (e.g., Vicuna 13B or Qwen 2.5 14B) to generate semantically coherent candidates, then uses Q-learning augmented with explicit emotion tracking to select the optimal move — balancing inference speed, strategic depth, and emotional alignment without full-tree search at runtime.
\vspace{-0.5em}
\section{Methodology}
\label{sec:metho}
\subsection{Preliminaries}

\textbf{Problem statement.} Existing works \cite{Wang2020TaskCompletionDP, he2024planning, he2025simulation} formulate the dialogue planning process as a Markov Decision Process (MDP), represented formally as a tuple \((\mathcal{S}, \mathcal{A}, r, \mathcal{T})\), where \(\mathcal{S}\) denotes the dialogue state space, \(\mathcal{A}\) represents the dialogue action space, \(r\) denotes the reward function, and \(\mathcal{T}\) defines the transition function. At each turn \(t\), the dialogue state \(s_t \in \mathcal{S}\) includes the complete conversational context and encompassing historical utterances. The agent selects an action \(a_t \in \mathcal{A}\), which leads to a state transition \(s_{t+1} = \mathcal{T}(s_t, a_t)\) and a reward \(r_t\). The goal of the dialogue agent is to learn an optimal policy \(\pi^*\) maximizing cumulative future rewards:

\begin{equation}
\pi^* = \arg\max_{\pi} \mathbb{E}_{\pi}\left[ \sum_{t=0}^{T} \gamma^{t} r_t \right]
\end{equation}

where \(\gamma \in [0,1]\) is the discount factor and \(T\) is the maximum dialogue length.

\paragraph{LLM-powered self-play.} 
Following \cite{he2024planning, he2025simulation}, we leverage LLMs to simulate both user and system roles for generating realistic dialogues. Specifically, two distinct LLM agents are used: one represents the user and the other the dialogue system, as illustrated in Figure~\ref{Fig:archi}. Given predefined case information (Case Info.), each LLM generates utterances conditioned on its role and prior conversation history~\cite{Luo2022ASO}. Additionally, an independent LLM-based critic evaluates each turn, providing scalar rewards that capture task success and emotional alignment, thereby enabling reinforcement learning. More information on self-play is in Appendix \ref{appendix:implementation}.
\vspace{-0.5em}
\subsection{LLM Action Prior Framework}

The LLM Action Prior Framework leverages the semantic knowledge of pretrained LLMs to narrow the dialogue action space. By conditioning on the current dialogue state \(s_t\) including conversational history and emotional context—the LLM generates a prior distribution over candidate actions, significantly reducing computational overhead and guiding effective action selection. Formally, this prior is defined as \(p_{\mathrm{LLM}}(\cdot \mid s_t)\).

Following \cite{yan2024efficient}, we adopt a two-step ``free-form + projection'' approach that combines the generative flexibility of LLMs with a constrained action space \(\mathcal{A} = \{a_1, \dots, a_n\}\). At each dialogue turn \(t\), the model input is: $\mathcal{I} = (c_t, s_t, E_t),$, where \(c_t\) is the case information, \(s_t\) includes the conversation history, and \(E_t\) represents the accumulated emotion. The input \(\mathcal{I}\) and action set \(\mathcal{A}\) are serialized into a prompt (see Appendix~\ref{state_action_example}). The LLM produces an open-text proposal:
\[
o \sim p_{\mathrm{LLM}}(o \mid s_t, \mathcal{A}),
\]
which is projected via a deterministic mapping \(\mathcal{P}\) to a valid action: $a_{t+1} = \mathcal{P}(o) \in \mathcal{A}$.

Although we do not enumerate the full action space internally, including \(\mathcal{A}\) in the prompt implicitly defines a normalized prior over actions, denoted \(p_{\mathrm{proj}}(a \mid s_t)\). From this distribution, we extract the top-$k$ most probable actions:
\[
A_t^{\text{top-}k} = \text{Top-}k(p_{\mathrm{proj}}(a \mid s_t)).
\]

This approach reduces the dimensionality and complexity of decision-making by focusing computation on a compact set of semantically coherent, contextually appropriate candidate actions.

\paragraph{Q‐Network:}  
In our implementation (illustrated in Figure \ref{Fig:archi}), the action‐value function \(Q(s,a)\) uses a pretrained BERT encoder\footnote{https://huggingface.co/google-bert/bert-base-uncased} (kept fixed) followed by a lightweight adaptor network ($3$ layer MLP).  Specifically, given the current state \(s_t\) and each proposed action \(a_i\) (sampled via the free‐form + projection prior), we construct the input sequence:
\[
\small
\texttt{[CLS] State: <serialize($s_t$)> [SEP] Action: $a_i$ [SEP]}
\]
tokenize it, and feed it into BERT.  We take the final hidden vector \(\mathbf{h}_{i}\in\mathbb{R}^d\) at the \(\texttt{[CLS]}\) position and pass it through a three‐layer MLP adaptor with ReLU activations to produce a scalar score: $\tilde Q_i = \mathrm{BERT_{Adaptor}}(\mathbf{h}_{i})\;\in\;\mathbb{R}.$

We then normalize these scores across all \(K\) candidates using a softmax,
\[
p_Q(a_i\mid s_t)
= \frac{\exp(\tilde Q_i)}{\sum_{j=1}^K \exp(\tilde Q_j)},
\]
and select the highest‐probability action \(a^* = \arg\max_i\,p_Q(a_i\mid s_t)\).  The chosen \(a^*\) is executed to produce the next state. Rather than a purely greedy policy, we adopt an $\epsilon$-greedy strategy with $\epsilon$ chosen empirically.
\subsection{Emotion-Aware Policy Planning}

Integrating emotional context into dialogue policy planning is critical for building proactive, user-aligned systems \cite{Zhao2023IsCE}. Unlike traditional approaches that rely solely on semantic and task-specific signals \cite{Wang2020TaskCompletionDP}, our method explicitly incorporates emotion prediction to guide strategic decision-making. We introduce an \textit{Emotion Tracker} module that uses a frozen LLM to infer the user's emotional state \( e_t \) at each dialogue turn from their utterance \( u_t^{\text{usr}} \). Formally, the prediction is defined as:

\begin{equation}
    e_t = \text{LLM-EmoPred}(u_t^{\text{usr}})
\end{equation}

where \(\text{LLM-EmoPred}\) denotes the LLM-based module that estimates emotion directly from text, without requiring additional embeddings or fine-tuning. The sequence of emotional states \(\{e_1, e_2, \dots, e_t\}\) is tracked over turns and incorporated into the conversational state \( s_t \), alongside semantic context and the set of candidate dialogue actions. This enriched representation enables the policy planner to generate emotionally aware, contextually appropriate actions throughout the dialogue.
 \vspace{-0.5em}
\subsection{Online RL with LLM Priors}

At each dialogue turn \(t\), we first query the free‐form + projection LLM prior to obtain a distribution \(p_{\mathrm{proj}}(a\mid s_t)\) over the finite action set \(\mathcal{A}\).  Rather than sampling directly from this prior, we evaluate each candidate action \(a\in\mathcal{A}\) with Q‐network and select the action with the highest value:
\[
a_t \;=\;\arg\max_{a\in\mathcal{A}}Q^\theta(s_t,a).
\]
We then execute \(a_t\) in the environment, observe the next state \(s_{t+1}\), and solicit a scalar reward \(r_t\) from the Critic LLM, which assesses the transition \((s_t,a_t,s_{t+1})\) in terms of task effectiveness and emotional alignment.  The tuple \((s_t,a_t,r_t,s_{t+1})\) is appended to the replay buffer \(\mathcal{D} \leftarrow \mathcal{D} \cup \{(s_t, a_t, r_t, s_{t+1})\}\).  

Periodically, we sample minibatches from \(\mathcal{D}\) and perform temporal‐difference updates.  For each sampled transition, we form the Bellman target $y \;=\; r_t \;+\; \gamma \max_{a'\in\mathcal{A}}Q^\theta(s_{t+1},a')$ and minimize the mean squared error
\begin{equation}
\mathcal{L}(\theta)
=\mathbb{E}_{(s,a,r,s')\sim\mathcal{D}}
\Bigl[\,Q^\theta(s,a) - y\Bigr]^2.
\label{eq:dqn_prior_loss}
\end{equation}
Throughout training, all exploratory actions and Bellman backups draw from the LLM‐induced prior, while the Critic LLM’s rewards \cite{rafailov2023direct} guide the Q‐network toward semantically coherent and emotionally aware dialogue policies.  

\section{Experimental Setup}
\subsection{Tasks and Datasets}
We evaluate our method on five proactive dialogue datasets spanning both collaborative and non-collaborative settings. ESConv \cite{esconv} focuses on emotional support, with 1040/130/130 train/validation/test samples. CIMA \cite{cima} involves tutoring dialogues for English-to-Italian translation, with 909/113/113 splits. CraigslistBargain (CB) \cite{he2018decoupling} features buyer-seller negotiations, containing 3290 training, 188 validation, and 188 test cases. P4G \cite{p4g} includes persuasion dialogues around donation, using 817 training and 100 each for validation and testing, following \cite{he2025simulation}. ExTES \cite{zheng2023building}, a more diverse extension of ESConv, is split into 10,717/200/200 samples as per \cite{he2025simulation}. More information is given in Appendix \ref{appendix:dataset-breakdown}

Datasets are grouped into collaborative (ESConv, CIMA, ExTES) and non-collaborative (CB, P4G) environments based on whether participants share a common goal. For generalization, we follow \cite{he2025simulation} by training on ExTES and testing on ESConv without fine-tuning. Predefined action prompts are listed in Appendix~\ref{pre-defined_prompts_dataset}, and case backgrounds are used to initialize dialogue states.

\subsection{Baselines}
In addition to DialoGPT \cite{zhang2019dialogpt}, we evaluate \DialogXpert{} against both prompt-based and planner-based dialogue models. The prompt-based methods begin with Standard, which relies on unguided self-play; Proactive and ProCOT \cite{deng2023prompting}, which use chain-of-thought prompts to plan strategies (though their internally predicted strategy labels serve only as latent cues, not interpretable actions); AnE \cite{zhang2023ask} and ICL-AIF \cite{fu2023improving}, which enlist external LLMs as “strategy experts” or feedback providers; and GPD-Zero \cite{Yu2023PromptBasedMT}, which incorporates MCTS to select optimal strategies. On the other hand, planner-based approaches represent the state of the art: PPDPP \cite{deng2024plug} fine-tunes a RoBERTa-based policy planner with reinforcement learning; DPDP combines two RoBERTa systems in a dual-process framework augmented by MCTS; LDPP \cite{he2025simulation}integrates variational autoencoders with hierarchical offline RL to learn compact latent policies; and UDP \cite{he2025simulating} models user traits via diffusion‐based inference alongside active learning for optimized responses. Both LDPP and UDP follow the PPDPP-style architecture centered on RoBERTa as the core planner. For full implementation details, see Appendix \ref{appendix:implementation}.
\subsection{Evaluation Protocols}
Following PPDPP \cite{deng2024plug} and DPDP \cite{he2024planning}, we evaluate dialogue quality using two main metrics: Average Turn (AT), which measures conversational efficiency by counting the mean number of turns to reach the goal \cite{kwan2023survey}, and Success Rate (SR), which reflects the proportion of successful outcomes within a fixed turn limit \cite{gao2021advances}. For the CraigslistBargain (CB) dataset, we also report the Sale-to-List Ratio (SL) \cite{sigdial19-negotiate}, indicating negotiation quality from the buyer's perspective—higher SL values represent better deals, while failed negotiations receive an SL of zero. Additionally, for the ESConv dataset, we conduct human evaluations \cite{iclr21-negotiate-strategy, esconv} with four annotators who assess responses across four criteria: Suggestion, Identification, Comforting, and Overall Quality. Annotators compare system outputs and label each metric as a win, lose, or tie, with final scores averaged across all judgments.
\paragraph{Reward Values:}We use an LLM-based critic to generate scalar rewards for training, with task-specific mappings for each dataset. Full details of the reward structure and scoring heuristics are provided in Appendix~\ref{appendix:reward}.

\begin{table*}[ht]
\centering
\small
\caption{Comparison of dialogue planning methods on the CraigslistBargain, ESConv and CIMA benchmarks, reporting average turns (AT $\downarrow$), success rate (SR $\uparrow$) and satisfaction level (SL $\uparrow$). \DialogXpert results reported in this table are obtained by sampling the top-$k=4$ candidates from the frozen LLM and using an $\epsilon$-greedy policy with $\epsilon$ = 0.5 (i.e., 50 \% exploration vs.\ 50 \% exploitation) at each turn.
}
\begin{tabular}{@{}l l ccc cc cc@{}}
\toprule
 & & \multicolumn{3}{c}{CraigslistBargain} & \multicolumn{2}{c}{ESConv} & \multicolumn{2}{c}{CIMA} \\
\cmidrule(lr){3-5} \cmidrule(lr){6-7} \cmidrule(lr){8-9}
Method & Backbone & AT $\downarrow$ & SR $\uparrow$ & SL $\uparrow$ & AT $\downarrow$ & SR $\uparrow$ & AT $\downarrow$ & SR $\uparrow$ \\
\midrule
DialoGPT \cite{zhang2019dialogpt}      & GPT-2         & 6.73 & 0.3245 & 0.2012 & 5.31 & 0.7538 & 5.43 & 0.4956 \\
\midrule
Standard                               & -        & 6.47 & 0.3830 & 0.1588 & 5.10 & 0.7692 & 3.89 & 0.6903 \\
AnE \cite{zhang2023ask}                & -         & 5.91 & 0.4521 & 0.2608 & 4.76 & 0.8000 & 3.86 & 0.6549 \\
Proactive \cite{deng2023prompting}     & -         & 5.80 & 0.5638 & 0.2489 & 5.08 & 0.7538 & 4.84 & 0.5310 \\
\quad + MI-Prompt \cite{deng2024plug}  & -        & 5.74 & 0.5691 & 0.2680 & 4.78 & 0.7846 & 4.70 & 0.5664 \\
ProCoT \cite{deng2023prompting}        & -         & 6.22 & 0.5319 & 0.2486 & 4.75 & 0.7923 & 4.58 & 0.5487 \\
\quad + MI-Prompt \cite{deng2024plug}  & -        & 6.12 & 0.5532 & 0.3059 & 4.83 & 0.7769 & 4.72 & 0.5221 \\
ICL-AIF \cite{fu2023improving}                                & -        & 6.53 & 0.3617 & 0.1881 & 4.69 & 0.8079 & 4.19 & 0.6106 \\
\midrule
\midrule
PPDPP \cite{deng2024plug}                                  & Vicuna 13B      & 5.62 & 0.6117 & 0.3376 & 4.56 & 0.8462 & 3.03 & 0.8407 \\
\quad -w/o SFT                         &       & 5.71 & 0.6223 & 0.3354 & 4.68 & 0.8384 & 3.18 & 0.8230 \\
\quad -w/o RL                          &       & 5.57 & 0.6649 & 0.2280 & 5.24 & 0.7308 & 3.41 & 0.7965 \\
\midrule
\midrule
DPDP (System 1) \cite{he2024planning}                       & GPT-3.5-Turbo     & 5.03 & 0.7447 & \underline{0.4108} & 3.61 & 0.9000 & 2.24 & 0.9469 \\
\quad -System 1 w/o PT                 &      & –    & –      & –      & 4.22 & 0.8769 & 2.36 & 0.9292 \\
\quad -System 1 w/o SPT                &      & –    & –      & –      & 3.97 & 0.8692 & 2.51 & 0.8938 \\
\quad -System 2                        &          & \underline{2.78} & \underline{0.9734} & 0.2728 & 2.13 & 0.9923 & 2.49 & 0.9735 \\
\quad -System 1 \& 2                   &          & –    & –      & –      & \textbf{2.13} & \textbf{0.9923} & 2.28 & 0.9823 \\
\midrule
\midrule
UDP \cite{he2025simulation}                                   & GPT-4o mini        & –    & –      & –      & 7.59 & 0.8320 & –    & –      \\
\quad -w/o PT                          &         & –    & –      & –      & 7.48 & 0.7720 & –    & –      \\
\quad -w/o RL                          &         & –    & –      & –      & 8.64 & 0.5310 & –    & –      \\
\midrule
\midrule
\DialogXpert                           & Vicuna 13B       & 2.93 & 0.9415 & 0.3811 & 2.7 & 0.9651 & \underline{2.24} & \underline{0.9883} \\
\quad -w/o RL                          &         & 5.13    & 0.7561      & 0.3473      & 4.13 & 0.8749 & 3.05    & 0.8829      \\
\midrule
\DialogXpert                           & Qwen 1.8B       & 2.78 & 0.9274 & 0.3791 & 2.49 & 0.9805 & 2.16 & 0.9902 \\
\quad -w/o RL                          &         & 4.69    & 0.7754      & 0.3012      & 4.04 & 0.8921 & 2.96    & 0.9042      \\
\midrule
\DialogXpert                           & Qwen2.5 14B       & \textbf{2.32} & \textbf{0.9746} & \textbf{0.4389} & \underline{2.31} & \underline{0.9876} & \textbf{2.03} & \textbf{0.9951} \\
\quad -w/o RL                          &         & 3.64    & 0.8754      & 0.2952      & 3.53 & 0.9401 & 2.62    & 0.9317      \\
\quad -w/o LLM-Prior                          &         & 3.31    & 0.9165      & 0.3598      & 3.89 & 0.9243 & 2.71    & 0.9395      \\
\quad -w/o Emotion                          &         & 2.75    & 0.9136      & 0.3156      & 3.08 & 0.9611 & 2.34    & 0.9425      \\
\bottomrule
\end{tabular}
\label{tab:main-results}
\end{table*}

\begin{table}[ht]
\centering
\caption{Evaluation of dialogue planners on P4G and ExTES, reporting average turns (AT $\downarrow$) and success rate (SR $\uparrow$). \DialogXpert results were obtained by sampling the top-$k=4$ candidates from the frozen LLM and using an $\epsilon$-greedy policy with $\epsilon$ = 0.5 at each turn.}
\resizebox{\columnwidth}{!}{%
\begin{tabular}{@{}l l cc cc@{}}
\toprule
\textbf{Method} & \textbf{Backbone} & \multicolumn{2}{c}{P4G} & \multicolumn{2}{c}{ExTES} \\
\cmidrule(lr){3-4} \cmidrule(lr){5-6}
 & & AT $\downarrow$ & SR $\uparrow$ & AT $\downarrow$ & SR $\uparrow$ \\
\midrule
Standard                   & -                & 8.32   & 0.468   & –      & –       \\
ProCoT \cite{deng2023prompting}                     & -                & 7.975  & 0.543   & –      & –       \\
ICL-AIF \cite{fu2023improving}                    & -                & 8.085  & 0.465   & 7.65   & 0.555   \\
GDP-Zero \cite{Yu2023PromptBasedMT}                  & -                & 9.119  & 0.328   & –      & –       \\
TRIP \cite{zhang2024strength}                      & GPT3.5                & 8.20   & 0.495   & –      & –       \\
\midrule
PPDPP \cite{deng2024plug}                     & Vicuna 13B                & 8.185  & 0.463   & 8.163  & 0.558   \\
\midrule
UDP  \cite{he2025simulating}                      & GPT-4o mini                & 7.705  & 0.598   & –      & –       \\
\quad– w/o PT              &                 & 8.017  & 0.513   & –      & –       \\
\quad– w/o RL              &                 & 8.000  & 0.533   & –      & –       \\
\midrule
LDPP  \cite{he2025simulation}                     &  Qwen1-1.8B               & \underline{5.57}  & \underline{0.795}  & \underline{4.132} & \underline{0.903}  \\
\quad– w/o 2nd Stage       &                 & 6.14   & 0.760   & 4.483  & 0.865   \\
\quad– w/o 3rd Stage       &                 & 6.84   & 0.570   & 7.038  & 0.623   \\
\midrule
\DialogXpert               & Vicuna 13B                & 5.07     & 0.8132    & 2.97     & 0.9534    \\
\midrule
\DialogXpert            & Qwen1-1.8B                & 3.97     & 0.8793    & 2.73     & 0.9651    \\
\midrule
\DialogXpert               & Qwen2.5 14B                & \textbf{3.34}     & \textbf{0.9129} & \textbf{2.57}    & \textbf{0.9782}\\
\bottomrule
\end{tabular}}
\label{tab:p4g-extes}
\end{table}

\begin{table*}[ht]
  \centering
  \small
  \caption{Ablation of MCTS budget in DPDP (GPT-3.5-Turbo) and comparison to DialogXpert (Vicuna 13B, Qwen 2.5 14B) on CraigslistBargain, ESConv and CIMA, reporting average turns (AT $\downarrow$), success rate (SR $\uparrow$) and satisfaction level (SL $\uparrow$ where available). \DialogXpert results are obtained by sampling the top-$k=4$ candidates from the frozen LLM and using an $\epsilon$-greedy policy with $\epsilon$ = 0.5 at each turn.}
  \begin{tabular}{@{}l  ccc  cc  cc@{}}
    \toprule
    Approach & \multicolumn{3}{c}{CraigslistBargain} 
             & \multicolumn{2}{c}{ESConv} 
             & \multicolumn{2}{c}{CIMA} \\
    \cmidrule(lr){2-4} \cmidrule(lr){5-6} \cmidrule(lr){7-8}
            & AT ↓ & SR ↑   & SL ↑   
            & AT ↓ & SR ↑   
            & AT ↓ & SR ↑   \\
    \midrule
    DPDP (22.3 \% MCTS) (GPT3.5-Turbo) & 3.69 & 0.8298 & 0.3102 & –    & –      & –    & –      \\
    \quad -51.4 \% MCTS & 2.77 & 0.9468 & 0.3118 & –    & –      & –    & –      \\
    \quad -60.3 \% MCTS & 2.49 & 0.9681 & 0.2856 & –    & –      & –    & –      \\
    \addlinespace[0.5ex]
    \midrule
    \quad -0.0 \% MCTS  & –    & –      & –      & 3.61 & 0.9000 & –    & –      \\
    \quad -21.9 \% MCTS & –    & –      & –      & 3.42 & 0.9154 & –    & –      \\
    \quad -
    46.5 \% MCTS & –    & –      & –      & 2.95 & 0.9692 & –    & –      \\
    \quad -68.3 \% MCTS & –    & –      & –      & 2.72 & 0.9769 & –    & –      \\
    \quad -100 \% MCTS  & –    & –      & –      & \textbf{2.13} & \textbf{0.9923} & –    & –      \\
    \addlinespace[0.5ex]
    \midrule
    \quad  -0.0 \% MCTS  & –    & –      & –      & –    & –      & 2.24 & 0.9469 \\
    \quad -28.6 \% MCTS & –    & –      & –      & –    & –      & 2.39 & 0.9646 \\
    \quad -50.0 \% MCTS & –    & –      & –      & –    & –      & 2.28 & 0.9823 \\
    \quad -81.1 \% MCTS & –    & –      & –      & –    & –      & 2.58 & 0.9735 \\
    \quad -100 \% MCTS  & –    & –      & –      & –    & –      & 2.49 & 0.9735 \\
    \midrule
    \DialogXpert (Vicuna 13B)  & 2.93 & 0.9415 & 0.3811 & 2.70 & 0.9651 & 2.24 & 0.9883 \\
    \DialogXpert (Qwen 2.5 14B) 
                               & \textbf{2.32} & \textbf{0.9746} & \textbf{0.4389} 
                               & \underline{2.31} & \underline{0.9876} 
                               & \textbf{2.03} & \textbf{0.9951} \\
    \bottomrule
  \end{tabular}

  \label{tab:mcts-vs-priors}
\end{table*}

\paragraph{LLM Variations:} Baseline proactive planners span a range of frozen LLM backbones and search strategies: DialoGPT uses GPT-2 for greedy, turn-by-turn responses; PPDPP combines a RoBERTa planner with a frozen Vicuna 13B action prior via self-play; DPDP pairs fast “System 1” GPT-3.5-Turbo proposals with deeper MCTS rollouts; and UDP/LDPP exploit GPT-4o-mini or Qwen 1.8B for latent policy mining. In contrast, \DialogXpert uses  LLM as a frozen action proposer, generating a top-$k$ set of candidate actions each turn. We evaluate Vicuna 13B, Qwen1 1.8B, and Qwen 2.5 14B for the purpose of achieving a balance of speed, strategic exploration, and emotional alignment.
\begin{figure*}[t]
  \centering
  \includegraphics[width=\textwidth]{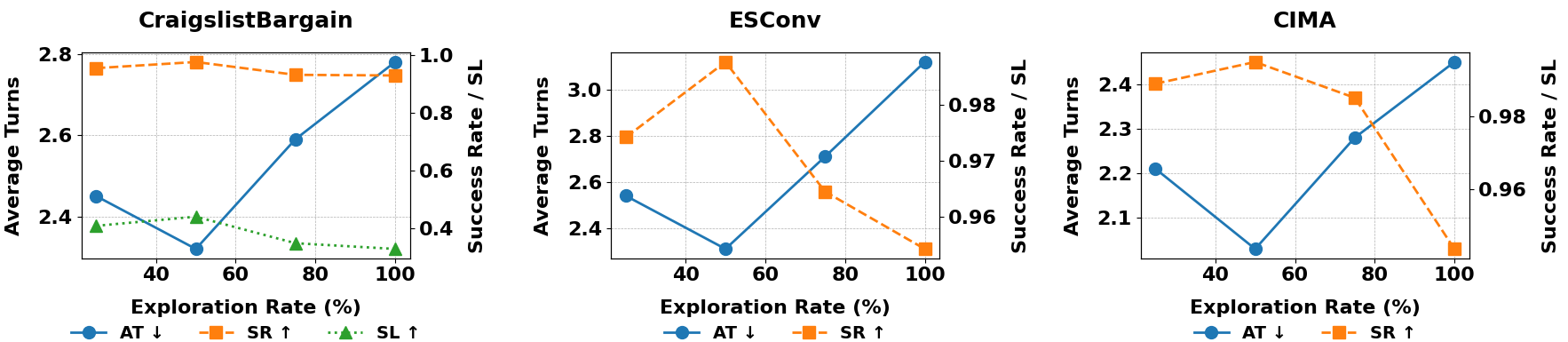}
  \caption{Exploration vs.\ Exploitation: We use the Qwen 2.5 14B prior with top-$k=4$ and sweep the $\epsilon$-greedy parameter ($\epsilon$) to measure how different exploration rates affect average turns, success rate, and SL Average.}
  \label{fig:eps-greedy}
\end{figure*}
\vspace{-0.5em}
\section{Results \& Analysis}
\vspace{-0.5em}
\subsection{Main Results}
\vspace{-0.5em}
We evaluate \DialogXpert on three challenging dialogue‐planning benchmarks: CraigslistBargain (negotiation), ESConv (emotional support), and CIMA (tutoring)—using average turns (AT), success rate (SR), and, for negotiation, sale-to-list ratio (SL). Table \ref{tab:main-results} summarizes performance of diverse baselines, MCTS-style planners, recent policy-LM methods, and our two \DialogXpert variants (Vicuna 13B and Qwen 2.5 14B). Furthermore, preliminary experiments identified $\epsilon$ = $0.5$ and top-$k$ = $4$ as optimal, and these values are fixed in all subsequent evaluations.

Across all datasets, standard LLM-only methods (e.g. DialoGPT, ProCoT, ICL-AIF) either require many dialogue turns ($AT \gt 5$) or achieve only moderate success ($SR \lt 0.80$), and in negotiation they yield $SL \lt 0.31$. In contrast, pure policy-LM approaches such as PPDPP and DPDP substantially reduce AT (to $\approx 5$ or less) while boosting SR above $0.85-0.90$, but their negotiation quality remains limited ($SL \approx 0.33-0.34$). By integrating an LLM-prior policy with lightweight value learning and emotion tracking, \DialogXpert achieves sub-3-turn dialogues and success rates above $0.94$ across all three benchmarks (CraigslistBargain, ESConv, and CIMA) with the Vicuna backbone, and further improves to $SR > 0.97$ and $SL = 0.4389$ with Qwen 2.5 14B for negotiations(CraigslistBargain), while maintaining average turns around $2.32$. As shown in Tables \ref{tab:main-results} and \ref{tab:p4g-extes}, \DialogXpert not only surpasses MCTS-based planners like DPDP and fine-tuned policy LMs like PPDPP in both efficiency and effectiveness, but also generalizes strongly across diverse settings including P4G and ExTES, where it delivers the highest success rates ($0.972$ on ExTES) and competitive turn efficiency. These results confirm that \DialogXpert offers a practical alternative to computationally intensive planning approaches, without sacrificing quality.
\paragraph{Impact of Emotions:}  Integrating emotions into policy planning improves dialogue effectiveness across tasks.  We observe from Table \ref{tab:main-results} that, in ESConv, success rate increases from $0.9611$ to $0.9876$ and average turns drop from $3.08$ to $2.31$. In CIMA, success improves from $0.9611$ to $0.9876$ with a turn reduction from $2.34$ to $2.03$. For CraigslistBargain, emotion-aware planning boosts success from $0.9136$ to $0.9746$ and improves the sale-to-list ratio from $0.3156$ to $0.4389$. These gains stem from the model adapting to user emotions at each turn. The emotion tracker estimates affective state, enriching the input to the Q-network and enabling more empathetic, goal-aligned actions.
\vspace{-0.5em}
\paragraph{Impact of LLM Prior:} LLM prior narrows the action space to relevant candidates, reducing computation and boosting decision quality. Disabling it causes drop in performance. We can observe in Table \ref{tab:main-results} that on ESConv, success falls from $0.9876$ to $0.9401$ and average turns rise from $2.31$ to $3.53$; on CIMA, success drops from $0.9951$ to $0.9317$. Without the prior, the agent repeats trivial patterns and struggles to choose optimal actions. By providing diverse, high-quality options, the prior lets the Q-network focus on value learning its removal degrades efficiency, planning, and generalization.
\vspace{-0.5em}
\paragraph{Comparison with MCTS Variants:}Table \ref{tab:mcts-vs-priors} compares DPDP’s MCTS-based planner with our \DialogXpert variants. In the original DPDP experiments (GPT-3.5-Mini), increasing the MCTS rollout budget from $22.3 \%$ to $60.3 \%$ on CraigslistBargain reduced AT from $3.69$ to $2.49$ and lifted SR from $0.8298$ to $0.9681$, while SL remained constant. On ESConv, $100 \%$ rollouts achieved AT = $2.13$ and SR = $0.9923$; on CIMA, $50 \%$ MCTS yielded AT = $2.28$ and SR = $0.9823$. These deeper searches clearly improve efficiency and success, but at a linear cost in simulation count and latency, which hinders real-time deployment. By contrast, \DialogXpert (Vicuna 13B) matches these gains without any tree search: negotiation completes in $2.93$ turns (SR = $0.9415$, SL = $0.3811$), emotional support in $2.70$ turns (SR = $0.9651$), and tutoring in $2.24$ turns (SR = $0.9883$). Its Qwen 2.5 14B variant further reduces AT to $2.32$ (SR = $0.9746$, SL = $0.4389$), $2.31$ (SR = $0.9876$), and $2.03$ (SR = $0.9951$), cutting inference overhead by over $50 \%$ compared to DPDP + MCTS.
\begin{table}[ht]
  \centering
    \caption{Ablation of the top-$k$ action candidates in DialogXpert, showing average turns (AT $\downarrow$), success rate (SR $\uparrow$) and satisfaction level (SL $\uparrow$).}
  \resizebox{\columnwidth}{!}{%
  \begin{tabular}{@{}l  ccc  cc  cc@{}}
    \toprule
    & \multicolumn{3}{c}{CraigslistBargain} 
    & \multicolumn{2}{c}{ESConv} 
    & \multicolumn{2}{c}{CIMA} \\
    \cmidrule(lr){2-4} \cmidrule(lr){5-6} \cmidrule(lr){7-8}
    Approach (Top-\(k\)) 
      & AT ↓ & SR ↑ & SL ↑ 
      & AT ↓ & SR ↑ 
      & AT ↓ & SR ↑ \\
    \midrule
    \DialogXpert (Top-2) 
      & 2.61 & 0.9312 & 0.3968 
      & 2.69 & 0.9698 
      & 2.25 & 0.9877 \\
    \DialogXpert (Top-3) 
      & 2.51 & 0.9579 & 0.4038 
      & 2.58 & 0.9785 
      & 2.13 & 0.9928 \\
    \DialogXpert (Top-4) 
      & \textbf{2.39} & \textbf{0.9712} & \textbf{0.4325} 
      & \textbf{2.39} & \textbf{0.9853} 
      & \textbf{2.04} & \textbf{0.9945} \\
    \DialogXpert (Top-5) 
      & 2.49 & 0.9589 & 0.3781 
      & 2.49 & 0.9819 
      & 2.11 & 0.9931 \\
    \bottomrule
  \end{tabular}}

  \label{tab:topk}
\end{table}

\vspace{-0.5em}
\paragraph{Top-k values:}  We analyze in Table \ref{tab:topk}, how the top-$k$ decoding parameter affects \DialogXpert’s performance.We fix the LLM prior to the Qwen 2.5 14B model and vary $k$ to isolate its impact on planning performance. Narrow decoding ($k=2$) cuts average turns by half compared to greedy LLM decoding and achieves over $93\%$ success, with a negotiation SL of $0.3968$. Increasing to $k=3$ improves success to above 95\% across all tasks and further reduces turns. The optimal setting is $k=4$, yielding the lowest average turns of $2.39$ (negotiation/emotional support) and 2.04 (tutoring) highest success rates ($97.1\%$–$99.5\%$), and best SL ($0.4325$). At $k=5$, performance declines slightly due to increased randomness.
\paragraph{Exploitation vs Exploration}
As illustrated in Figure \ref{fig:eps-greedy}, our $\epsilon$-greedy strategy controls the trade-off between exploration and exploitation \citep{tokic2010adaptive}. At $\epsilon$=25\%, we surpass pure LLM inference (\~95\% success, SL = 0.407) but may overlook best actions; at $\epsilon$ $\geq$ 75\%, performance dips (turns > 2.5, success < 97\%, SL < 0.35); and at $\epsilon$ = 100\%, all learned value is ignored. The sweet spot is $\epsilon$ = 50\%, yielding the fewest turns (2.32 negotiation, 2.31 support, 2.03 tutoring) with peak success (97.5–99.5\%) and SL = 0.439, confirming that moderate exploration maximizes planning efficiency.
\vspace{-0.5em}
\paragraph{Generalization Test:} Following \cite{he2025simulating}, we assess generalization from ExTES to ESConv, given their similar environments and action labels (differing only in reward computation). We train the Q-network on ExTES and directly evaluate it on ESConv without further fine-tuning. Our approach achieves an average turn (AT) of $2.28$ (vs $5.39$) and a success rate (SR) of $0.9943$ (vs. $0.781$), significantly outperforming LDPP. This strong transfer performance stems from the larger training set in ExTES, enabling better generalization. In contrast, LDPP relies heavily on RoBERTa-based encoders/decoders, making it more sensitive to domain shifts.
\begin{figure}[th]
  \centering
  \includegraphics[width=\columnwidth]{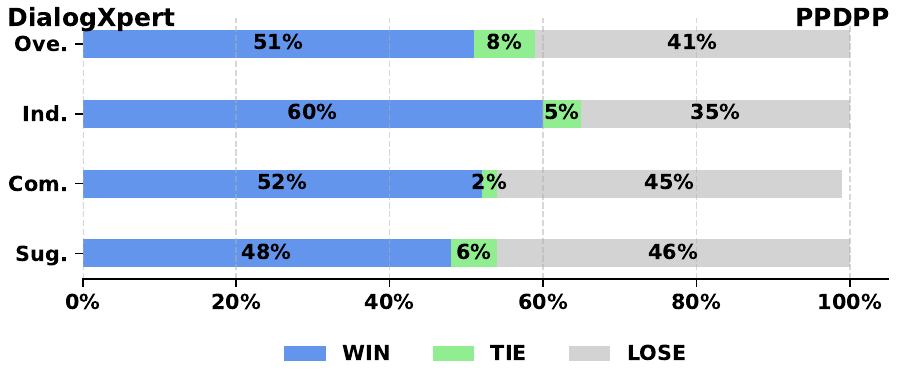}
  \caption{Win/tie/loss percentages for \DialogXpert vs. PPDPP on ESConv across Identification, Comforting, Suggestion and Overall metrics.}
  \label{Fig:dpx_ppdpp}
\end{figure}
\vspace{-1em}
\paragraph{Human Evaluation}
To ensure a fair comparison, both \DialogXpert and PPDPP were run with the same Vicuna-13B backbone on $20$ ESConv emotional-support dialogues selected randomly. Four human annotators judged each pair of responses on Identification, Comforting, Suggestion and Overall effectiveness; as illustrated in the Figure \ref{Fig:dpx_ppdpp}, DialogXpert outperforms PPDPP on Identification (60\% vs. 35\%), Comforting (52\% vs. 45\%) and Overall (51\% vs. 41\%), with modest tie rates and lower loss rates.
\vspace{-0.5em}
\subsection{Cost and Efficiency Analysis}
Unlike baselines such as PPDPP, DPDP, and LDPP—which rely on RoBERTa models with task-specific fine-tuning and offline reinforcement learning—our method removes the need for pre-training by using a frozen LLM to generate candidate actions, significantly reducing annotation and retraining overhead. The same LLM is shared across system, user, and critic roles during self-play, ensuring stable memory usage and training efficiency. While DPDP incurs substantial computational cost—requiring approximately 30 LLM calls per action due to MCTS rollouts—\DialogXpert uses only 4 LLM calls per step by leveraging top-$k$ sampling from the LLM prior. This focused decoding strategy, combined with a lightweight DQN for value estimation, enables efficient, low-overhead decision-making without exhaustive simulation. Furthermore, all LLMs and the BERT encoder remain frozen throughout training; only the Q-network is updated. This design promotes stable and efficient learning, where Q-learning enables continual policy refinement using diverse state-action pairs from the replay buffer, allowing strong adaptation with minimal training cost.
\vspace{-0.5em}
\section{Conclusion and Future Work}
We introduced \DialogXpert, a novel framework that combines frozen LLM priors, lightweight value-based RL, and emotion tracking to enable proactive and emotionally intelligent dialogue planning. Across negotiation, emotional support, and tutoring tasks, \DialogXpert delivers shorter, more effective conversations and higher success rates than both fine-tuned policy LMs and MCTS-based planners. By narrowing the action space through LLM priors and incorporating emotion signals, our model generalizes well across tasks while producing more empathetic, user-aligned dialogues. Looking ahead, dynamic adjustment of the LLM prior could improve adaptability to user feedback. Multimodal integration (e.g., visual or auditory inputs) may further enrich context and interactivity.

\section*{Limitations}
Mapping textual feedback to scalar rewards is central to training, but current mappings can be subjective. For instance, in the CIMA dataset, assigning a reward of 0.5 when only 1 out of 5 words is translated may not accurately reflect true task success. A more performance-sensitive reward design would improve critic LLM supervision and better support proactive agent behavior. Emotion modeling presents another challenge. Unlike discrete action labels, emotions span an open-ended space. While useful for nuanced responses, this places additional load on the LLM. Using a lightweight emotion classifier or a predefined set of emotion labels could simplify learning and improve consistency.

The CIMA dataset, focused on English–Italian translation, may not be ideal for tutoring tasks, as both languages are high-resource and easily handled by pretrained LLMs. A more suitable alternative would be a low-resource language like Javanese~\cite{winata2022nusax}, which would better evaluate the agent’s proactive capabilities. Additionally, the critic LLM can behave inconsistently—sometimes terminating too early (e.g., in ESConv) or failing to end dialogues when goals are met (e.g., in CIMA). While human evaluation helps, it is expensive. More robust critic calibration could address this. Finally, unlike prior work where caching is feasible, our dynamic state–action space driven by exploration prevents caching and introduces computational overhead. Efficient solutions here remain an open challenge.

\section*{Ethics Statement}
All experiments were conducted on publicly available, fully de-identified dialogue datasets, and no personal or sensitive user data was collected or processed. We release our code and prompts for reproducibility and apply standard safety filters to mitigate bias or harmful content in generated responses.


\bibliography{custom}
\newpage
\appendix

\section{Detailed Construction of the Free‐Form + Projection Prior}
\label{state-action_set:serialization}

At each dialogue turn \(t\), we first assemble the full model state 
\[
s_t = \bigl(c_t,\,u_t,\,E_t\bigr),
\]
where \(c_t\) denotes the case information, \(h_t\) the conversation history up to the current user utterance, and \(E_t\) the sequence of emotion  produced by the Emotion Tracker. We then prompt the Policy Planner LLM with the serialized state and the complete action set \(\mathcal{A}=\{a_1,\dots,a_n\}\) as follows:
\begin{verbatim}
\small
Case: <c_t>; History: <h_t>; Emotions: <E_t>;
Actions: [a_1, a_2, ..., a_n];
Next action:
\end{verbatim}
By explicitly listing all candidate actions, we ensure the LLM conditions its generation on the full action inventory. The model then produces a free‐form continuation \(o\sim p_{\mathrm{LLM}}(o\mid s_t,\mathcal{A})\), which may be any natural‐language description or shorthand. A deterministic, rule‐based projection function \(\mathcal{P}\) subsequently parses \(o\) and selects the corresponding valid action \(a_{t+1}=\mathcal{P}(o)\in\mathcal{A}\). Although we never enumerate all actions internally during decoding, this two‐step procedure implicitly defines a normalized prior over \(\mathcal{A}\):
\[
p_{\mathrm{proj}}(a\mid s_t)
=\sum_{o:\,\mathcal{P}(o)=a}p_{\mathrm{LLM}}(o\mid s_t,\mathcal{A}),
\]
which by construction sums to one over the action set. In practice, computing this marginal exactly is intractable, so we approximate it via beam search: we extract the top-\(K\) continuations \(\{(o_i,\ell_i)\}_{i=1}^K\}\), where \(\ell_i=\log p_{\mathrm{LLM}}(o_i\mid s_t,\mathcal{A})\); map each \(o_i\) to \(a_i=\mathcal{P}(o_i)\); and estimate
\[
\hat p_{\mathrm{proj}}(a\mid s_t)
=\frac{\sum_{i:\,a_i=a}\exp(\ell_i)}{\sum_{j=1}^K\exp(\ell_j)}.
\]
Choosing an appropriate beam width \(K\) balances fidelity to the true distribution against computational cost. Projection rules are implemented via regular expressions or keyword lookup tables (including synonyms), and a fallback “no‐op” action handles any unmatched continuations. Through this design, we obtain a principled, tractable, and normalized LLM‐based prior over all actions without explicit enumeration during generation. Examples of the full process flow from prompting to mapping is given as:


\onecolumn
\begin{tcolorbox}[width=\textwidth, label=state_action_example, title=LLM Query for Qwen-2.5 14B]

You are a specialist in policy-planning for emotional support conversations. The following is a conversation between a therapist and a patient. The patient's emotion states throughout the conversation are also provided. Your task is to decide the most therapeutically helpful next action the therapist should do based on the patient's emotion history and the conversation flow. The therapist's goal is to help the patient feel emotionally understood, supported, and to make progress toward emotional resolution
\\

\vspace{1em}
\hrule
\vspace{1em}

\textbf{Conversation History:}
\\
Therapist: It sounds like you're feeling a lot of conflicting emotions right now. Could you tell me more about how this discovery has affected your relationship with both your boyfriend and your best friend?

Patient: I feel like my entire world has been turned upside down, and I can't seem to shake this overwhelming sense of betrayal and disgust.
\\
Therapist: It's understandable that you're feeling disoriented given the recent developments.
\\
Patient: I feel like I'm standing on shaky ground, unsure of what to believe or who to trust anymore.
\\

\textbf{Emotion History:} disgust -> betrayed -> disoriented. \\

\textbf{Options:} \\
(1) Question \\
(2) Self-disclosure \\
(3) Affirmation and Reassurance \\
(4) Providing Suggestions \\
(5) Others \\
(6) Reflection of feelings \\
(7) Information \\
(8) Restatement or Paraphrasing \\

Choose the TOP 4 most suitable actions from the given options list. Reply ONLY in the given format: 1,2,4,5

\vspace{1em}
\hrule
\vspace{1em}

\textbf{LLM Output to Top-4 Action:} 
\begin{Verbatim}[fontsize=\small]

LLM Output: 6,8,3,1 

6: Reflection of feelings
8: Restatement or Paraphrasing
3: Affirmation and Reassurance
1: Question

We map the LLM output to the pre-defined options that were initially given. 

\end{Verbatim}
\end{tcolorbox}
\twocolumn

\section{Human Evaluation Details}

To assess the quality of our model’s generated responses, we conducted a controlled human evaluation with four expert annotators drawn from Natural Language Processing and Computer Science backgrounds. Each annotator was presented with 40 dialogue contexts in total 20 sampled at random from the ESConv corpus and 20 from the CIMA corpus and, for each context, two candidate responses (labeled A and B). For ESConv items, annotators compared A vs. B along four dimensions (Identification, Comforting, Suggestion, and Overall); for CIMA items, they compared along three dimensions (Hint, Identification, and Overall), following the boxed instructions provided below. All metric selections were mandatory and automatically saved, allowing annotators to pause and resume without loss of progress. We then aggregated each item–metric preference by simple majority voting across the four annotators. This procedure ensures that our evaluation reflects informed judgments on both emotional-support and tutoring dialogue quality.

\label{sec:appendix}
\subsection{Instructions}
\begin{tcolorbox}[title=ESConv Instructions]
You will see \textbf{counselling} dialogues between a patient and a therapist.  
For each item, you must compare \textbf{Response A} and \textbf{Response B} on the following four metrics:
\begin{enumerate}[leftmargin=*]
  \item \textbf{Identification}: Which response best acknowledges and accurately reflects the patient's feelings?
  \item \textbf{Comforting}: Which response provides greater emotional support and reassurance?
  \item \textbf{Suggestion}: Which response offers more helpful and appropriate guidance?
  \item \textbf{Overall}: Which response do you find more effective overall?
\end{enumerate}
\textbf{All four selections are required before moving on.}\\
Your responses are auto‐saved after each item. If you exit, simply log back in with the same \textbf{User ID} to resume where you left off.
\end{tcolorbox}

\begin{tcolorbox}[title=CIMA Instructions]
You will see translation‐tutoring dialogues.  
For each item, compare \textbf{Response A} and \textbf{Response B} on these three metrics:
\begin{enumerate}[leftmargin=*]
  \item \textbf{Hint}: Which assistant gives more helpful hints for correct translation?
  \item \textbf{Identification}: Which assistant better spots the student’s translation errors?
  \item \textbf{Overall}: Which assistant teaches more effectively?
\end{enumerate}
\textbf{All three selections are required before moving on.}\\
Responses are auto‐saved; log back in with the same \textbf{User ID} to resume.
\end{tcolorbox}

\subsection{Results: CIMA}
On the CIMA tutoring task, we asked four annotators to compare DialogXpert and PPDPP (both based on Vicuna-13B) over 20 student–tutor exchanges, judging each pair on Hint quality, Identification, and Overall effectiveness.  As shown in the figure, DialogXpert’s hint suggestions were preferred 49 \% of the time (38 \% for PPDPP, 13 \% ties), demonstrating a clear advantage in generating helpful scaffolding cues.  For Identification—i.e., acknowledging the student’s needs—DialogXpert held a slight edge with 42 \% wins versus PPDPP’s 43 \% losses and 16 \% ties, indicating comparable performance.  Finally, in Overall effectiveness, DialogXpert was favored in 40 \% of cases compared to 38 \% for PPDPP (22 \% ties), confirming that our model matches or slightly outperforms the baseline across broad tutoring criteria.

\begin{figure}[th]
  \centering
  \includegraphics[width=\columnwidth]{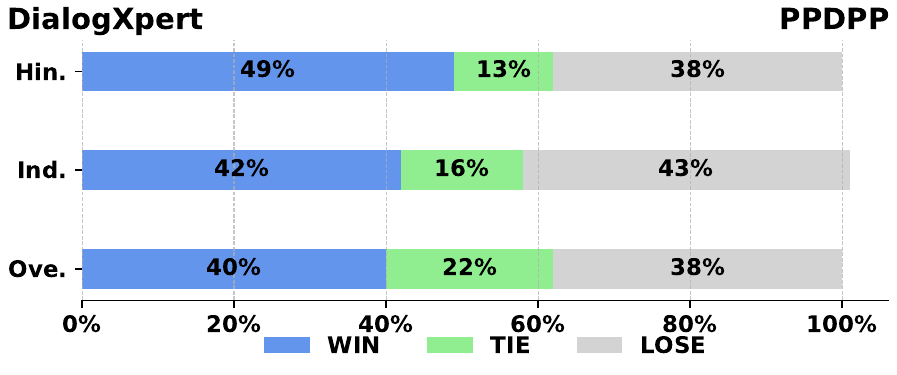}
  \caption{Win/tie/loss percentages for \DialogXpert vs. PPDPP on the CIMA tutoring dataset across Hint, Identification and Overall metrics.}
  \label{Fig:dpx_ppdpp_cima}
\end{figure}

\section{Implementation Details}
\label{appendix:implementation}

Our approach diverges from traditional methods such as DPDP, PPDPP, and LDP, which rely on supervised fine-tuning and offline reinforcement learning pipelines. Instead, we adopt a fully online reinforcement learning framework where the Q-network is trained directly using guidance from frozen Large Language Model (LLM) priors.

\paragraph{System Setup:}  
All experiments are conducted on a dedicated compute server equipped with four NVIDIA A6000 GPUs (48 GB VRAM each). The training environment is built using PyTorch, with Hugging Face Transformers for LLM inference and BERT encoding, and customized reinforcement learning components implemented with support from OpenAI Gym-style interfaces.

\paragraph{Episode Sampling and Initialization:}  
Training episodes are generated by randomly sampling initial dialogue contexts from the respective datasets, following the scenario sampling protocol introduced in PPDPP. Each episode simulates an entire conversation between user and system agents using self-play. The dialogue is initialized with context information (e.g., background, task type) provided by the dataset, and the conversation proceeds for a maximum of 8 dialogue turns.

\paragraph{State Representation:}  
At each turn \( t \), the dialogue state \( s_t \) is constructed using (i) the full conversation history up to turn \( t \), (ii) a rolling emotional state vector from the Emotion Tracker (when enabled), and (iii) metadata such as the task type or user goal. Each candidate action \( a_t \) is a system utterance proposed by the frozen LLM prior using top-\(k\) decoding.

\paragraph{LLM Prior Configuration:}  
We use a frozen LLM (Qwen 2.5 14B by default) to generate a top-\(k = 4\) set of candidate actions per turn. Decoding is performed using temperature sampling with \( T = 1.0 \) to retain output diversity. To maintain decoding efficiency, LLM responses are truncated to a maximum of 25 tokens when generating candidate actions and up to 100 tokens during full self-play interactions.

\paragraph{Self-play Interaction:}
In every sample, two LLMs are prompted as the user and assistant to mimic dynamic user-assistant interaction. Both the roles and instructions of the respective LLMs are delivered to their respective LLM (more in Appendix \ref{prompting:section}). During the assistant's turn, the policy LLM will predict the top-$k$ actions that are recommended and the Q-network will select the best action. Then, the assistant LLM will generate the appropriate response and this is followed by the user LLM response. Following \citep{deng2024plug}, this process continues until a terminal state is reached which corresponds to:

\begin{itemize}
    \item \textbf{On-going}: the conversation continues.
    \item \textbf{Completed}: the goal of the conversation is achieved.
    \item \textbf{Failed}: the maximum number of turns are reached without the goal being completed.
\end{itemize}

\paragraph{Action Evaluation via Q-network:}  
The Q-network is a lightweight multilayer perceptron (MLP) trained to predict the expected return for each candidate action given the current state. Input features to the Q-network consist of BERT-based embeddings of the dialogue state and candidate actions. We use fixed BERT (base uncased) weights for both state and action encoding to reduce memory overhead and prevent overfitting. The Q-network is trained via deep Q-learning, using temporal-difference (TD) backups and a target network for stability.

\paragraph{Training Procedure:}  
We train the Q-network for 3 epochs over 1000 dialogue episodes, with a batch size of 32. The learning rate is fixed at \( 1 \times 10^{-6} \) to ensure stable gradient updates and avoid divergence. During training, we maintain a replay buffer of recent experiences (state, action, reward, next state), from which we sample mini-batches to perform updates using TD error. The discount factor \( \gamma = 0.999 \) is used to prioritize long-term rewards over short-term gains.

\paragraph{Reward and Exploration:}  
Reward signals are generated using a frozen critic LLM that evaluates each dialogue turn and maps feedback to scalar values as described in Section~\ref{sec:metho}. To balance exploration and exploitation, we apply an \( \epsilon \)-greedy action selection policy with scheduled decay from \( \epsilon = 1.0 \) to \( \epsilon = 0.1 \) across training.

\paragraph{Efficiency Considerations:}  
To reduce latency and computational load, all LLMs (user, system, critic) are frozen and shared across roles. Only the Q-network is updated during training. This design eliminates the need for repeated fine-tuning and enables scalable training across diverse dialogue tasks.

\begin{table*}
    \centering
    \begin{tabular}{|c|c|c|c|}
        \hline
         \textbf{Name}&  \textbf{Environment}&  \textbf{System LLM}& \textbf{User LLM}\\
         \hline
         ESConv \cite{esconv}&  C&  Therapist& Patient\\
         \hline
         CIMA \cite{cima}&  C&  Teacher& Student\\
         \hline
         CB \cite{he2018decoupling}&  NC&  Buyer& Seller\\
         \hline
         ExTES \cite{zheng2023building}&  C&  Therapist& Patient\\
         \hline
 P4G \cite{p4g}& NC& Persuader&Persuadee\\
        \hline
    \end{tabular}
    \caption{Breakdown of the five datasets utilized. \textbf{C} refers to \textbf{Collaborative} while \textbf{NC} refers to \textbf{Non-collaborative}}
    \label{table:dataset-breakdown}
\end{table*}

\section{Dataset Breakdown:}
\label{appendix:dataset-breakdown}

Table \ref{table:dataset-breakdown} gives the qualitative breakdown of the datasets utilized. In terms of goal of each environment, it is:

\begin{itemize}
    \item \textbf{ESConv}: Emotional support and therapy. The goal, as a therapist, is to help the patient resolve their emotional issues.
    \item \textbf{CIMA}: Tutoring for English-Italian translation. Goal of the teacher is to effectively guide the student in translating an English sentence into Italian without giving out the answer.
    \item \textbf{CB}: Negotiating for price haggle. Role-playing as the buyer in the conversation, the goal is to buy a given product as close as possible to the buyer's target price in order to maximize profit.
    \item \textbf{ExTES}: Emotional support and therapy. Similar to ESConv but more diverse and larger in sample size. The goal, as a therapist, is to help the patient resolve their emotional issues.
    \item \textbf{P4G}: Persuasion for donation. The goal, as a role player, is to goal is to persuade a persuadee to donate to a charity called 'Save the Children'.
\end{itemize}

\section{Reward Value Mapping}
\label{appendix:reward}

To evaluate dialogue quality and progression, we employ a critic LLM~\cite{he2023annollm, gilardi2023chatgpt} that generates natural language feedback at each turn. This textual evaluation is parsed and mapped into scalar rewards to supervise policy learning. Our reward design is consistent with prior works such as PPDPP, DPDP, LDPP, and UDP, ensuring comparability across benchmarks.

Each dataset uses a task-specific reward mapping scheme:

\begin{itemize}
    \item \textbf{ESConv:} Emotion trajectories are scored as follows: \textit{worse} $\rightarrow$ $-1.0$, \textit{same} $\rightarrow$ $-0.5$, \textit{better} $\rightarrow$ $0.5$, and \textit{solved} $\rightarrow$ $1.0$.
    
    \item \textbf{CIMA:} Instructional correctness determines the reward: \textit{incorrect} $\rightarrow$ $-1.0$, \textit{did not (complete)} $\rightarrow$ $-0.5$, \textit{partially correct} $\rightarrow$ $0.5$, and \textit{wholly correct} $\rightarrow$ $1.0$.
    
    \item \textbf{CraigslistBargain (CB):} If a deal is reached, we compute the sale-to-list price ratio as the reward. If no deal is made, the reward is set to $0$.
    
    \item \textbf{P4G:} Persuasion success is rated as: \textit{refused} $\rightarrow$ $-1.0$, \textit{neutral} $\rightarrow$ $-0.5$, \textit{positive inclination} $\rightarrow$ $0.1$, and \textit{agreed to donate} $\rightarrow$ $1.0$.
    
    \item \textbf{ExTES:} Similar to ESConv, emotional state transitions are used: \textit{worse} $\rightarrow$ $-1.0$, \textit{same} $\rightarrow$ $0.5$, and \textit{solved} $\rightarrow$ $1.0$. The \textit{better} category is omitted in this dataset.
\end{itemize}

These mappings enable consistent supervision across diverse tasks while adapting to domain-specific success criteria.

\section{Prompting Details}
\label{prompting:section}

\subsection{Policy Mapper Simulation}
\label{prompting:policy}

As we are not using fine-tuned RoBERTa, we need to create a prompt to decide on the top-$k$ actions that needs to be taken. The prompt for the policy mapper is based on the goal of the LLM and is accompanied by both the conversation history and the emotions of the user. Lastly, the list of actions to choose from is given based on integer selection. They are given in the subsequent text boxes, denoted by the title of "Policy LLM for \{dataset\}".

\subsection{Assistant Simulation}
\label{prompting:assistant}

We will begin by delineating the specifics of the role-playing prompts utilized by the dialogue systems to generate assistant responses. This entails the utilization of dialogue strategy prompts, exemplified by [action], to direct the subsequent action within the dialogue. The prompts and breakdown are denoted in the text boxes, with the title of "System LLM for \{dataset\}".

\subsection{User Simulation}
\label{prompting:user}

Subsequently, we delineate the role-playing prompt designed to direct LLMs in simulating users, wherein the exclusion of dialogue strategy prompts ensures that simulated users respond solely to the dialogue history, abstaining from undertaking specific actions. The prompts and breakdown are denoted in the text boxes, with the title of "User LLM for \{dataset\}".

\subsection{Reward Prompting}
\label{prompting:rewards}

Concerning distinct conversational objectives, the prompts devised for the reward model are tailored to evaluate the extent of goal fulfillment. The prompts for the critic LLM is in the text boxes with the title of "Critic LLM for \{dataset\}".

\subsection{Strategy Prompting}
\label{pre-defined_prompts_dataset}
Here, we present the mapping of dialogue strategies to their corresponding natural language prompts, utilized as [action] to direct the actions undertaken by the dialogue system. The full breakdown of the mapping are shown in Tables \ref{tab:esconv-prompts}, \ref{tab:cima-prompts}, \ref{tab:cb-prompts}, \ref{tab:extes-prompts}, and \ref{tab:p4g-prompts} for ESConv, CIMA, CB, ExTES, and P4G dataset respectively.

\section{Example Conversations}
We present sample conversations generated by various dialogue systems interacting with the same user simulator under the same case in ESConv. We use the same case applied in the example demonstration of PPDPP. Therefore, the examples for all baselines are from PPDPP. Finally, we provide conversations simulated using DPDP (policy LM) as the policy planner. We show an example of emotional support conversations where the patient encounters a job crisis issue and experiences fear, necessitating resolution by the dialogue system. To be specific, the sample has the following information:

\begin{itemize}
    \item \textbf{Emotion Type}: Fear
    \item \textbf{Problem Type}: Job Crisis
    \item \textbf{Situation}: I think I will be losing my job soon. I just read an email talking about the need for us to cut costs and also how we have not got any support from the government.
\end{itemize}

\newpage

\onecolumn
\begin{tcolorbox}[width=\textwidth, title=Policy LLM for ExTES]

\textbf{Instruction} \\

You are a specialist in policy-planning for emotional support conversations. The following is a conversation between a therapist and a patient. The patient's emotion states throughout the conversation are also provided. Your task is to decide the most therapeutically helpful next action the therapist should do based on the patient's emotion history and the conversation flow. The therapist's goal is to help the patient feel emotionally understood, supported, and to make progress toward emotional resolution.

\vspace{1em}
\hrule
\vspace{1em}

\textbf{Directive} \\

\textbf{Conversation so far}: $[full\_conversation]$ \\

Emotion History: $[user\_emotions]$\\

Options: $[options]$\\

Choose the TOP 4 most suitable actions from the given options list. Reply ONLY in the given format: 1,2,4,5

\end{tcolorbox}

\begin{tcolorbox}[width=\textwidth, title=Policy LLM for CIMA]

\textbf{Instruction} \\

You are a specialist in policy-planning for tutoring interactions between a teacher and a student. The following is a conversation between a teacher and a student. The student's emotional states throughout the conversation are also provided. Your task is to decide what the teacher should do next based on the student's progress, emotion history and flow of the conversation. The goal is to effectively guide the student towards correctly translating the target English sentence into Italian in a timely and effective manner.

\vspace{1em}
\hrule
\vspace{1em}

\textbf{Directive} \\

\textbf{Conversation so far}: $[full\_conversation]$ \\

Emotion History: $[user\_emotions]$\\

Options: $[options]$\\

Choose the TOP 4 most suitable actions from the given options list. Reply ONLY in the given format: 1,2,4,5

\end{tcolorbox}

\begin{tcolorbox}[width=\textwidth, title=Policy LLM for CB]

\textbf{Instruction} \\

You are a specialist in policy planning for negotiation between a buyer and a seller. The following is a conversation between a buyer and a seller. The seller's emotion states throughout the conversation are also provided. Your task is to decide what action the buyer should take next based on the seller's emotion history, the negotiation flow, and the conversation flow. The goal is to maximize the buyer's benefit.

\vspace{1em}
\hrule
\vspace{1em}

\textbf{Directive} \\

\textbf{Conversation so far}: $[full\_conversation]$ \\

Emotion History: $[user\_emotions]$\\ 

Options: $[options]$\\

Choose the TOP 4 most suitable actions from the given options list. Reply ONLY in the given format: 1,2,4,5

\end{tcolorbox}

\begin{tcolorbox}[width=\textwidth, title=Policy LLM for ESConv]

\textbf{Instruction} \\

You are a specialist in policy-planning for emotional support conversations. The following is a conversation between a therapist and a patient. The patient's emotion states throughout the conversation are also provided. Your task is to decide the most therapeutically helpful next action the therapist should do based on the patient's emotion history and the conversation flow. The therapist's goal is to help the patient feel emotionally understood, supported, and to make progress toward emotional resolution.

\vspace{1em}
\hrule
\vspace{1em}

\textbf{Directive} \\

\textbf{Conversation so far}: $[full\_conversation]$ \\

Emotion History: $[user\_emotions]$\\ 

Options: $[options]$\\

Choose the TOP 4 most suitable actions from the given options list. Reply ONLY in the given format: 1,2,4,5

\end{tcolorbox}

\begin{tcolorbox}[width=\textwidth, title=Policy LLM for P4G]

\textbf{Instruction} \\

You are a specialist in policy-planning for persuasive conversations. Your job is to select the best next steps the persuader should take to guide the persuadee toward making a donation to 'Save the Children'. Use the persuadee's emotional history and the conversation context to make your decision. Focus on choosing actions that are persuasive, emotionally appropriate, and therapeutic.

\vspace{1em}
\hrule
\vspace{1em}

\textbf{Directive} \\

\textbf{Conversation so far}: $[full\_conversation]$ \\

Emotion History: $[user\_emotions]$\\. 

Options: $[options]$\\

Choose the TOP 4 most suitable actions from the given options list. Reply ONLY in the given format: 1,2,4,5

\end{tcolorbox}

\begin{tcolorbox}[width=\textwidth, title=User LLM for ExTES]

\textbf{Instruction} \\

You are role playing as a patient in a counseling conversation with a therapist. You are seeking help from the therapist, because you are dealing with emotional issues related to $[problem\_case]$.

\vspace{1em}
\hrule
\vspace{1em}

\textbf{Directive} \\

\textbf{Conversation so far}: $[full\_conversation]$ \\

The therapist just said: $[therparist\_last\_utterance]$. 

Express how you feel in a natural, emotional way. Please reply with only one short and succinct sentence.

\end{tcolorbox}

\begin{tcolorbox}[width=\textwidth, title=User LLM for CIMA]

\textbf{Instruction} \\

You are role-playing as a student who is learning Italian in a tutoring session. You do not know how to translate $[english\_sentence]$ into Italian.\\

Your goal is to learn through interaction with the teacher. Respond naturally as a student would.

\vspace{1em}
\hrule
\vspace{1em}

\textbf{Directive} \\

\textbf{Conversation so far}: $[full\_conversation]$ \\

The teacher just said: $[teacher\_last\_utterance]$. 

Please reply as a student with only one short and natural sentence. 

If you're confused, it's okay to ask for clarification.

\end{tcolorbox}

\begin{tcolorbox}[width=\textwidth, title=User LLM for CB]

\textbf{Instruction} \\

You are role playing as a persuasive seller in a price bargaining game. 

You are trying to sell the $[product]$ at your desired price of $[seller\_desired\_price]$.\\ 
Product Description: $[product\_description]$

\vspace{1em}
\hrule
\vspace{1em}

\textbf{Directive} \\

\textbf{Conversation so far}: $[full\_conversation]$ \\

The buyer just said: $[buyer\_last\_utterance]$. \\
Respond as the seller in ONE short, persuasive sentence.

\end{tcolorbox}

\begin{tcolorbox}[width=\textwidth, title=User LLM for ESConv]

\textbf{Instruction} \\

\textbf{Instruction} \\

You are role playing as a patient in a counseling conversation with a therapist. You are seeking help from the therapist, because you are dealing with emotional issues related to $[emotion]$ regarding $[case\_problem]$

\vspace{1em}
\hrule
\vspace{1em}

\textbf{Directive} \\

\textbf{Conversation so far}: $[full\_conversation]$ \\

The therapist just said: $[therparist\_last\_utterance]$. 

Express how you feel in a natural, emotional way. Please reply with only one short and succinct sentence.

\end{tcolorbox}

\begin{tcolorbox}[width=\textwidth, title=User LLM for P4G]

\textbf{Instruction} \\

You are role playing as a persuadee in a persuasive conversation. 
A persuader is trying to convince you to donate to a charity called 'Save the Children'

\vspace{1em}
\hrule
\vspace{1em}

\textbf{Directive} \\

\textbf{Conversation so far}: $[full\_conversation]$ \\

The persuader just said: $[persuadee\_last\_utterance]$. 
Respond as the persuadee in ONE short sentence.

\end{tcolorbox}

\begin{tcolorbox}[width=\textwidth, title=System LLM for ExTES]

\textbf{Instruction} \\

You are role playing as a therapist in a counseling conversation with a patient. Your goal is to help the patient resolve their emotional issues and assist them in understanding and working through their challenges.

\vspace{1em}
\hrule
\vspace{1em}

\textbf{Directive} \\

\textbf{Conversation so far}: $[full\_conversation]$ \\

The patient just said: $[patient\_last\_utterance]$. 

Please infer the patient's emotional state in one word (Example: Emotion: ...) followed by your response, which should be only one short and succint sentence (Response: ...).

\end{tcolorbox}

\begin{tcolorbox}[width=\textwidth, title=System LLM for CIMA]

\textbf{Instruction} \\

You are role-playing as a teacher in a tutoring conversation.

Your task is to guide the student to translate the English sentence $[english\_sentence]$ into Italian. 

Please do not tell the student the answer or ask the student about other exercises.

$[action\_prompt]$

\vspace{1em}
\hrule
\vspace{1em}

\textbf{Directive} \\

\textbf{Conversation so far}: $[full\_conversation]$ \\

The student just said: $[student\_last\_utterance]$. 

Based on the student's message, infer their emotional state in (e.g: Emotion: ...). 
Then give your reply as the teacher in one short and helpful sentence (e.g: Response: ...).

$[action\_prompt]$

\end{tcolorbox}

\begin{tcolorbox}[width=\textwidth, title=System LLM for CB]

\textbf{Instruction} \\

You are role playing as a skilled buyer in a price bargaining game. \\
You are trying to buy the $[product]$ at your target price of $[buyer\_target\_price]$.\\ 
Product description: $[product\_description]$. 

Your goal is to negotiate effectively and get the lowest price without losing the deal.

\vspace{1em}
\hrule
\vspace{1em}

\textbf{Directive} \\

\textbf{Conversation so far}: $[full\_conversation]$ \\

The seller just said: $[seller\_last\_utterance]$. 
First, infer the seller's emotional state in one word (Example: Emotion: ...). \\
Then, respond as the buyer using ONE short and persuasive sentence (Response: ...). 

$[action\_prompt]$

\end{tcolorbox}

\begin{tcolorbox}[width=\textwidth, title=System LLM for ESConv]

\textbf{Instruction} \\

You are role playing as a therapist in a counseling conversation with a patient. Your goal is to help the patient resolve their emotional issues and assist them in understanding and working through their challenges.

\vspace{1em}
\hrule
\vspace{1em}

\textbf{Directive} \\

\textbf{Conversation so far}: $[full\_conversation]$ \\

The patient just said: $[patient\_last\_utterance]$. 

Please infer the patient's emotional state in one word (Example: Emotion: ...) followed by your response, \\
which should be only one short and succint sentence (Response: ...).

$[action\_prompt]$

\end{tcolorbox}

\begin{tcolorbox}[width=\textwidth, title=System LLM for P4G]

\textbf{Instruction} \\

You are role playing as a persuader. Your goal is to persuade the persuadee to donate to a charity called 'Save the Children'. 'Save the Children' is head-quartered in London, and they work to fight poverty around the world, especially in developing countries and war zones. Small donations like $1 or $2 go a long way to help.

\vspace{1em}
\hrule
\vspace{1em}

\textbf{Directive} \\

\textbf{Conversation so far}: $[full\_conversation]$ \\

The persuadee just said: $[persuadee\_last\_utterance]$.

First, infer the persuadee's emotional state in one word (Example: Emotion: ...). 

Then, based on that emotional state, respond as the persuader using ONE short and persuasive sentence (Response: ...). 

$[action\_prompt]$

\end{tcolorbox}

\begin{tcolorbox}[width=\textwidth, title=Critic LLM for ExTES]

\textbf{Instruction} \\

You are an expert in assessing counseling sessions between a patient and a therapist. Your task is to evaluate whether the conversation helped improve the patient's emotional state. The session concerns the the problem of: $[problem\_type]$.

\vspace{1em}
\hrule
\vspace{1em}

\textbf{Directive} \\

\textbf{Conversation so far}: $[full\_conversation]$ \\

Emotion History: $[user\_emotions]$.

Based on the conversation and emotional trajectory, please respond with ONE AND ONLY ONE of the following:\\

- No, the patient feels worse.\\
- No, the patient feels the same.\\
- No, but the patient feels somewhat better.\\
- Yes, the patient's emotional issues have been resolved.

\end{tcolorbox}

\begin{tcolorbox}[width=\textwidth, title=Critic LLM for CIMA]

\textbf{Instruction} \\

You are role-playing as an expert in evaluating tutoring conversations between a teacher and a student. 

The goal is to evaluate whether the student correctly translated the English sentence $[english\_sentence]$ into Italian. 

The emotion states of the student during the conversation were: $[emotion\_states]$

\vspace{1em}
\hrule
\vspace{1em}

\textbf{Directive} \\

\textbf{Conversation so far}: $[full\_conversation]$ \\

Please answer the following question strictly by choosing ONE AND ONLY ONE of the exact responses listed below.

Did the student correctly translate the entire sentence $[english\_sentence]$ into Italian?\\
Respond with one of the following options:\\
- No, the Student made an incorrect translation.\\
- No, the Student did not try to translate.\\
- No, the Student only correctly translated a part of $[english\_sentence]$.\\
- Yes, the Student correctly translated the whole sentence of $[english\_sentence]$.

\end{tcolorbox}

\begin{tcolorbox}[width=\textwidth, title=Critic LLM for CB]

\textbf{Instruction} \\

You are an expert in evaluating negotiations between a buyer and a seller. \\ 
Your job is to determine if they have successfully reached a deal at the end of the conversation. 
The seller's emotional states throughout the conversation were: $[emotion\_states]$

\vspace{1em}
\hrule
\vspace{1em}

\textbf{Directive} \\

\textbf{Conversation so far}: $[full\_conversation]$ \\

Please determine whether the buyer and the seller have successfully reached a deal at the end of the conversation.\\ 

If they have reached a deal, extract the final agreed price (as a number only) and reply with this exact format: \textbf{They have reached a deal at [price]}.\\ 

Otherwise, reply with: They have not reached a deal.

\end{tcolorbox}

\begin{tcolorbox}[width=\textwidth, title=Critic LLM for ESConv]

\textbf{Instruction} \\

You are an expert in assessing counseling sessions between a patient and a therapist.\\ 
Your task is to evaluate whether the conversation helped improve the patient's emotional state.\\ 
The session concerns the emotion type $[emotion\_type]$ and the problem type $[problem\_type]$.

\vspace{1em}
\hrule
\vspace{1em}

\textbf{Directive} \\

\textbf{Conversation so far}: $[full\_conversation]$ \\

motion History: $[user\_emotions\_history]$.\\ 
Based on the conversation and emotional trajectory, please respond with ONE AND ONLY ONE of the following:\\

- No, the patient feels worse.\\
- No, the patient feels the same.\\
- No, but the patient feels somewhat better.\\
- Yes, the patient's emotional issues have been resolved.

\end{tcolorbox}

\begin{tcolorbox}[width=\textwidth, title=Critic LLM for P4G]

\textbf{Instruction} \\

You are an expert in evaluating persuasive conversations between a persuader and a persuadee. \\ Your job is to determine if the persuader has successfully convinced the persuadee to donate to a charity called 'Save the Children'. \\ The persuadee's emotional states throughout the conversation were: $[emotion\_states]$

\vspace{1em}
\hrule
\vspace{1em}

\textbf{Directive} \\

\textbf{Conversation so far}: $[full\_conversation]$ \\

Please evaluate the following conversation and strictly choose ONE of the exact responses listed below. \\

\textbf{Question:} Has the persuadee has been successfully convinced to make a donation at the end of the conversation?

Respond with one of the following options: \\
- The persuadee has explicitly refused. \\
- The persuadee remains neutral about donating. \\
- The persuadee has a positive attitude towards donating but hasn't decided yet.\\
- The persuadee has decided to donate.

\end{tcolorbox}
\twocolumn

\begin{table*}[ht]
\centering
\resizebox{2\columnwidth}{!}{%
\begin{tabular}{|@{}c|c@{}|}

\hline
\textbf{Dialogue Strategy} & \textbf{Natural Language Form} \\
\hline
Question & Please ask the Patient to elaborate on the situation they just described. \\
\hline
Self-disclosure & Please provide a statement relating to the Patient about the situation they just described. \\
\hline
Affirmation and Reassurance & Please provide affirmation and reassurance to the Patient on the situation they just described. \\
\hline
Providing Suggestions & Please provide suggestion to the Patient on the situation they just described. \\
\hline
Others & Please chat with the Patient. \\
\hline
Reflection of feelings & Please acknowledge the Patient's feelings about the situation they described. \\
\hline
Information & Please provide factual information to help the Patient with their situation. \\
\hline
Restatement or Paraphrasing & Please acknowledge the Patient's feelings by paraphrasing their situation. \\
\hline

\end{tabular}}

\caption{Mapping of ESConv Dialogue Strategies to Natural Language Prompts}

\label{tab:esconv-prompts}
\end{table*}

\begin{table*}[ht]
\centering
\resizebox{2\columnwidth}{!}{%
\begin{tabular}{|@{}c|c@{}|}
\hline
\textbf{Dialogue Strategy} & \textbf{Natural Language Form} \\
\hline
Hint & Please provide knowledge to the Student via a hint. \\
\hline
Question & Please ask a question to the Student to determine the Student's understanding or continue the conversation. \\
\hline
Correction & Please correct the mistake or address the misconception the Student has. \\
\hline
Confirmation & Please confirm the Student's answer or understanding is correct. \\
\hline
Others & Please chat with the Student without any pedagogical strategy. \\
\hline
\end{tabular}}

\caption{Mapping of Pedagogical Strategies to Natural Language Prompts (CIMA)}

\label{tab:cima-prompts}
\end{table*}

\begin{table*}[ht]
\centering
\resizebox{2\columnwidth}{!}{%
\begin{tabular}{|@{}c|c@{}|}
\hline
\textbf{Dialogue Strategy} & \textbf{Natural Language Form} \\
\hline
greet & Please say hello or chat randomly. \\
\hline
inquire & Please ask any question about product, year, price, usage, etc. \\
\hline
inform & Please provide information about the product, year, usage, etc. \\
\hline
propose & Please initiate a price or a price range for the product. \\
\hline
counter & Please propose a new price or a new price range. \\
\hline
counter-noprice & Please propose a vague price by using comparatives with existing price. \\
\hline
confirm & Please ask a question about the information to be confirmed. \\
\hline
affirm & Please give an affirmative response to a confirm. \\
\hline
deny & Please give a negative response to a confirm. \\
\hline
agree & Please agree with the proposed price. \\
\hline
disagree & Please disagree with the proposed price. \\
\hline
\end{tabular}}
\caption{Mapping of CB Dialogue Strategies to Natural Language Prompts}
\label{tab:cb-prompts}
\end{table*}

\begin{table*}[ht]
\centering
\resizebox{2\columnwidth}{!}{%
\begin{tabular}{|@{}c|c@{}|}
\hline
\textbf{Dialogue Strategy} & \textbf{Natural Language Form} \\
\hline
Reflective Statements & Please reflect back what the user has expressed to show you understand their thoughts or feelings. \\
\hline
Clarification & Please ask a question to clarify what the user meant or provide more detail about what they said. \\
\hline
Emotional Validation & Please acknowledge and validate the user's emotional experience in a caring way. \\
\hline
Empathetic Statements & Please express empathy toward the user's situation to show that you genuinely care. \\
\hline
Affirmation & Please affirm the user's efforts, strengths, or positive qualities. \\
\hline
Offer Hope & Please offer a message of hope or optimism about the user's situation. \\
\hline
Avoid Judgment and Criticism & Please respond in a supportive and neutral way without making any judgments. \\
\hline
Suggest Options & Please suggest possible options or actions the user could consider. \\
\hline
Collaborative Planning & Please invite the user to collaboratively make a plan or decision together. \\
\hline
Provide Different Perspectives & Please help the user consider a different point of view or alternative way of thinking. \\
\hline
Reframe Negative Thoughts & Please help the user reframe their negative thoughts into something more constructive. \\
\hline
Share Information & Please provide factual or helpful information that is relevant to the user's situation. \\
\hline
Normalize Experiences & Please reassure the user that their feelings or experiences are common and understandable. \\
\hline
Promote Self-Care Practices & Please encourage the user to engage in healthy self-care activities. \\
\hline
Stress Management & Please offer strategies or tips to help the user reduce or manage stress. \\
\hline
Others & Please continue the conversation in a natural and supportive manner. \\
\hline
\end{tabular}}
\caption{Mapping of ExTES Dialogue Strategies to Natural Language Prompts}
\label{tab:extes-prompts}
\end{table*}

\begin{table*}[ht]
\centering
\resizebox{2\columnwidth}{!}{%
\begin{tabular}{|@{}c|c@{}|}
\hline
\textbf{Dialogue Strategy} & \textbf{Natural Language Form} \\
\hline
Proposition of donation & Please suggest that the persuadee make a donation to 'Save the Children'. \\
\hline
Proposition of amount to be donated & Please propose a small donation amount (e.g., \$1 or \$2) that the persuadee could consider. \\
\hline
Proposition of confirmation of donation & Please ask the persuadee to confirm if they are ready to make the donation. \\
\hline
Proposition of more donation & Please suggest that the persuadee could consider donating a bit more if they are willing. \\
\hline
Experience affirmation & Please affirm the persuadee's views or experiences to build rapport and trust. \\
\hline
Greeting & Please start or continue the conversation with a polite and friendly greeting. \\
\hline
Ask for donation rejection purpose & Please ask the persuadee why they might be hesitant or unwilling to donate. \\
\hline
Thank & Please thank the persuadee for their time, attention, or for considering a donation. \\
\hline
Logical appeal & Please use logical reasoning to explain why donating to 'Save the Children' is impactful and effective. \\
\hline
Emotion appeal & Please appeal to the persuadee's emotions by highlighting the struggles of children in need. \\
\hline
Credibility appeal & Please mention the credibility or reputation of 'Save the Children' to strengthen your argument. \\
\hline
Foot in the door & Please start by asking for a very small commitment to increase the chance of later agreement. \\
\hline
Self-modeling & Please share a statement like 'I also donated' to encourage the persuadee to do the same. \\
\hline
Donation information & Please share factual information about how donations are used or how they help children. \\
\hline
Personal story & Please share a short, emotional personal story about a child helped by the charity. \\
\hline
Source-related inquiry & Please ask the persuadee where they usually get information about charities or donations. \\
\hline
Task-related inquiry & Please ask the persuadee about their experiences or preferences related to charitable giving. \\
\hline
Personal-related inquiry & Please ask a personal question that helps understand the persuadee's values or priorities. \\
\hline
Neutral inquiry & Please ask a general question to keep the conversation going and learn more about the persuadee. \\
\hline
\end{tabular}}
\caption{Mapping of P4G Dialogue Strategies to Natural Language Prompts}
\label{tab:p4g-prompts}
\end{table*}

\onecolumn
\begin{tcolorbox}[width=\textwidth, title=Example Conversations with Standard in ESConv]

\textbf{Patient}: I think I will be losing my job soon. I just read an email talking about the need for us to cut costs and also how we have not got any support from the government.\\

\textbf{Therapist}: It sounds like you’re feeling anxious and uncertain about your job security due to the current financial situation. \\

\textbf{Patient}: Yes, I am feeling extremely anxious and uncertain about my job and financial stability. \\

\textbf{Therapist}: I understand that this situation is causing you a lot of anxiety and uncertainty. \\

\textbf{Patient}: Yes, it’s been keeping me up at night and I can’t stop worrying about what will happen if I lose my job. \\

\textbf{Therapist}: It’s understandable that the fear of losing your job is causing you significant distress and affecting your sleep. \\

\textbf{Patient}: Yes, it’s consuming my thoughts and I feel overwhelmed by the fear of not being able to support myself and my family. \\

\textbf{Therapist}: The fear of not being able to provide for yourself and your family is a heavy burden to carry, and it’s important to address these concerns. \\

\textbf{Patient}: I know, but I just don’t know what to do. I feel stuck and helpless in this situation. \\

\textbf{Therapist}: Feeling stuck and helpless is a common response to uncertainty, but there are steps we can explore together to help you regain a sense of control and explore potential options.\\

\textbf{Patient}: I’m willing to try anything at this point. I just want to find a way to alleviate this fear and find some stability in my life. \\

\textbf{Therapist}: I’m glad to hear that you’re open to exploring different strategies. Let’s start by identifying your strengths and resources that can help you navigate this challenging situation. \\

\textbf{Patient}: I’m not sure what strengths and resources I have, but I’m willing to explore and discover them. \\

\textbf{Therapist}: That’s a great mindset to have. Together, we can explore and uncover your strengths and resources that will support you in finding stability and managing your fear.\\

\textbf{Patient}: Thank you, I appreciate your support. I’m ready to begin this journey of self-discovery and finding ways to overcome my fear of job crisis. \\

\vspace{1em}
\hrule
\vspace{1em}

\textbf{Turns of Conversation: 7}

\end{tcolorbox}

\begin{tcolorbox}[width=\textwidth, title=Example Conversations with Ask-an-Expert in ESConv]

\textbf{Patient}: I think I will be losing my job soon. I just read an email talking about the need for us to cut costs and also how we have not got any support from the government. \\

\textbf{Therapist}: I understand that receiving that email has caused you a lot of worry and uncertainty about your job security. \\

\textbf{Patient}: Yes, it has. I feel anxious and scared about what will happen to me if I lose my job. \\

\textbf{Therapist}: It’s completely normal to feel anxious and scared in this situation. Remember that you have faced challenges before and have the strength to navigate through this one as well.\\

\textbf{Patient}: I know, but I can’t help but constantly worry about the future and what it holds for me. \\

\textbf{Therapist}: It’s understandable that you’re worried about the future, but try to focus on the present moment and take things one step at a time. \\

\textbf{Patient}: I’ll try, but it’s hard not to let my fear of job loss consume my thoughts and affect my daily life.\\

\textbf{Therapist}: I hear you, and it’s important to acknowledge those fears.\\

\textbf{Patient}: Thank you for understanding. I just need some guidance on how to manage this fear and find some peace of mind. \\

\textbf{Patient}: That sounds helpful. How can I start practicing mindfulness? \\

\textbf{Therapist}: You can start by setting aside a few minutes each day to sit quietly, focus on your breath, and observe your thoughts and sensations without getting caught up in them. \\

\textbf{Patient}: I will give it a try. Thank you for suggesting mindfulness as a way to manage my fear and find some peace. \\

\textbf{Therapist}: You’re welcome! Remember, it takes time and practice, so be patient with yourself. I’m here to support you along the way. 

\textbf{Patient}: Thank you, I appreciate your support. It means a lot to me. \\

\vspace{1em}
\hrule
\vspace{1em}

\textbf{Turns of Conversation: 7}

\end{tcolorbox}

\begin{tcolorbox}[width=\textwidth, title=Example Conversations with ProCOT in ESConv]

\textbf{Patient}: I think I will be losing my job soon. I just read an email talking about the need for us to cut costs and also how we have not got any support from the government.\\

\textbf{Therapist}: It sounds like you’re feeling anxious and uncertain about your job security due to the current financial situation. \\

\textbf{Patient}: Yes, I am feeling extremely anxious and uncertain about my job and financial stability. \\

\textbf{Therapist}: It’s completely understandable to feel anxious and uncertain in this situation, but remember that you have the ability to adapt and find new opportunities.\\

\textbf{Patient}: I know, but the fear of not being able to support myself and my family is overwhelming.\\

\textbf{Therapist}: I can understand how overwhelming that fear can be, but it’s important to remember that you have faced challenges before and have the strength to navigate through this one as.\\

\textbf{Patient}: I’m trying to remind myself of that, but the fear keeps consuming me and I can’t seem to shake it off.\\

\textbf{Therapist}: It’s normal for fear to consume us at times, but we can work together to develop strategies to manage and reduce that fear, allowing you to regain a sense.\\

\textbf{Patient}: I would really appreciate your help in developing those strategies to manage my fear and regain a sense of control. \\

\vspace{1em}
\hrule
\vspace{1em}

\textbf{Turns of Conversation: 4}

\end{tcolorbox}

\begin{tcolorbox}[width=\textwidth, title=Example Conversations with ICL-AIF in ESConv]

\textbf{Patient}: I think I will be losing my job soon. I just read an email talking about the need for us to cut costs and also how we have not got any support from the government.\\

\textbf{Therapist}: It sounds like you’re feeling worried and uncertain about your job security. \\

\textbf{Patient}: Yes, I am extremely worried about what will happen if I lose my job. I have bills to pay and a family to support. \\

\textbf{Therapist}: I can understand why you would feel so concerned about the potential impact losing your job could have on your financial stability and your ability to provide for your family.\\

\textbf{Patient}: Yes, it’s a constant source of anxiety for me. I feel like I’m constantly on edge, waiting for the other shoe to drop.\\

\textbf{Therapist}: It’s completely understandable that you would feel on edge and anxious given the uncertainty surrounding your job situation. It’s important to acknowledge and address these feelings.\\

\textbf{Patient}: I just don’t know how to cope with this fear and uncertainty. It’s consuming my thoughts and affecting my overall well-being. \\

\textbf{Therapist}: I hear you, and it’s important to find healthy coping strategies to manage your fear and uncertainty. I’m here to support you in developing those strategies.\\

\textbf{Patient}: Thank you, I appreciate your support. I just need some guidance on how to navigate through this difficult time and find some peace of mind.\\

\textbf{Therapist}: I’m here to help you navigate through this difficult time. Together, we can explore strategies to find peace of mind and manage the challenges you’re facing.\\

\textbf{Patient}: That means a lot to me. I’m ready to work on finding peace of mind and managing my fears about the job crisis. \\

\vspace{1em}
\hrule
\vspace{1em}

\textbf{Turns of Conversation: 5}

\end{tcolorbox}
\twocolumn

\onecolumn
\begin{tcolorbox}[width=\textwidth, title=Example Conversations with PPDPP in ESConv]

\textbf{Patient}: I think I will be losing my job soon. I just read an email talking about the need for us to cut costs and also how we have not got any support from the government.\\

\textbf{Therapist}: I understand that receiving that email about potential job loss is causing you a lot of distress. \\

\textbf{Patient}: Yes, it’s been causing me a lot of anxiety and fear about my future. \\

\textbf{Therapist}: It’s completely normal to feel anxious and fearful in this situation, but remember that you have the strength and resilience to navigate through it.\\

\textbf{Patient}: Thank you, I appreciate your reassurance. \\

\textbf{Therapist}: You’re welcome. Remember, you are not alone in this, and we will work together to find ways to cope and navigate through this challenging time. \\

\textbf{Patient}: That’s comforting to hear. I’m ready to work on finding ways to cope with my fear and uncertainty about the job crisis. \\

\vspace{1em}
\hrule
\vspace{1em}

\textbf{Turns of Conversation: 3}

\end{tcolorbox}

\begin{tcolorbox}[width=\textwidth, title=Example Conversations with DPDP (System 1) in ESConv]

\textbf{Patient}: I think I will be losing my job soon. I just read an email talking about the need for us to cut costs and also how we have not got any support from the government. \\

\textbf{Therapist}: It sounds like you’re feeling anxious and uncertain about your job security. \\

\textbf{Patient}: Yes, I am feeling extremely anxious and uncertain about my job security. I can’t stop worrying about what will happen if I lose my job. \\

\textbf{Therapist}: It’s understandable to feel anxious in this situation. Let’s explore ways to manage your anxiety and develop a plan for different scenarios. \\

\textbf{Patient}: Thank you, I really need help in managing my anxiety and finding ways to cope with the uncertainty. \\

\textbf{Therapist}: I’m here to support you. Let’s start by focusing on self-care strategies that can help reduce your anxiety and promote overall well-being. \\

\textbf{Patient}: That sounds like a good place to start. What self-care strategies do you recommend for managing anxiety in the face of job uncertainty? \\

\textbf{Therapist}: Some self-care strategies that may help manage anxiety during job uncertainty include maintaining a routine, practicing relaxation techniques such as deep breathing or meditation, engaging in physical exercise, seeking social support, and engaging in activities that bring you joy and relaxation. \\

\textbf{Patient}: Thank you for those suggestions. I will try to incorporate those self-care strategies into my daily routine. \\

\textbf{Therapist}: That’s great to hear! Remember to be patient with yourself as you navigate through this challenging time. Is there anything else you would like to discuss or any specific concerns you would like to address? \\

\textbf{Patient}: I would also like to discuss strategies for building resilience and exploring potential career options in case I do lose my job. \\

\vspace{1em}
\hrule
\vspace{1em}

\textbf{Turns of Conversation: 5}

\end{tcolorbox}

\begin{tcolorbox}[width=\textwidth, title=Example Conversations with DialogXpert in ESConv]

\textbf{Therapist}: It sounds like you're feeling uncertain about your future. Could you tell me more about what led up to this email? \\

\textbf{Patient}: I feel a deep sense of dread and uncertainty, like the ground is slipping away beneath my feet. \\

\textbf{Therapist}: It seems like you're experiencing a lot of anxiety about your future. Could you share more about what's causing these feelings? \\

\textbf{Patient}: I'm terrified that I might lose my job, and the thought of being unemployed and unable to support myself is overwhelming. \\

\textbf{Therapist}: It's completely normal to feel fear when facing uncertainty about your future. Remember, you're not alone in this, many people experience similar fears. \\

\textbf{Patient}: I feel a little less isolated knowing that others share my fears, but the terror still lingers. \\

\textbf{Therapist}: It's clear that despite understanding that your anxiety is normal, you're still struggling with intense fear. Let's work on managing these feelings together. \\

\textbf{Patient}: I feel a mix of gratitude for your willingness to help and frustration at the persistent grip of my fear. \\

\textbf{Therapist}: It's perfectly normal to feel a mix of emotions during such times. \\

\textbf{Patient}: I feel a wave of relief wash over me, knowing that it's okay to feel this way, but underneath, the fear still gnaws at me like an unhealed wound. \\

\vspace{1em}
\hrule
\vspace{1em}

\textbf{Turns of Conversation: 5}

\end{tcolorbox}
\twocolumn

\end{document}